%% file: main.tex
\documentclass{article}

    \PassOptionsToPackage{numbers, compress}{natbib}
    \usepackage[final]{neurips_2023}

\usepackage[utf8]{inputenc} % allow utf-8 input
\usepackage[T1]{fontenc}    % use 8-bit T1 fonts
\usepackage{hyperref}       % hyperlinks
\usepackage{url}            % simple URL typesetting
\usepackage{booktabs}       % professional-quality tables
\usepackage{amsfonts}       % blackboard math symbols
\usepackage{nicefrac}       % compact symbols for 1/2, etc.
\usepackage{microtype}      % microtypography
\usepackage{xcolor}         % colors
\usepackage{shortcutsg}
\usepackage{graphicx}
\usepackage{booktabs}
\usepackage{hyperref}
\usepackage{url}
\usepackage{amsmath,amssymb,amsfonts}
\usepackage{graphicx}

\usepackage{wrapfig}
\usepackage{bbm}
\usepackage{footnote}
\makesavenoteenv{tabular}
\makesavenoteenv{table}
\usepackage{float}
\newcommand{\rep}{\phi}
\newcommand{\nwrep}{\varphi}
\usepackage{tikz}
\usepackage{booktabs}
\usetikzlibrary{arrows, automata, bayesnet, calc}
\usetikzlibrary{positioning}
\usepackage{subcaption}
\usepackage{soul}

\def\SPSB#1#2{\rlap{\textsuperscript{#1}}\SB{#2}}
\def\SP#1{\textsuperscript{#1}}
\def\SB#1{\textsubscript{#1}}
\newcommand{\nwb}{NW\SP{B}}

\newcommand{\nwbe}{NW\SPSB{B}{e}}
\newcommand{\ermb}{ERM\SP{B}}
\DeclareMathOperator{\supp}{supp}
\newcommand{\NW}{f}

\title{Learning Invariant Representations with a \\ Nonparametric Nadaraya-Watson Head}

\author{%
  Alan Q.~Wang \\%\thanks{
  Cornell University\\
  \And
  Minh Nguyen \\
  Cornell University \\
  \And
  Mert R. Sabuncu \\
  Cornell University \\
}

\begin{document}

\maketitle

\begin{abstract}
Machine learning models will often fail when deployed in an environment with a data distribution that is different than the training distribution. 
When multiple environments are available during training, many methods exist that learn representations which are invariant across the different distributions, with the hope that these representations will be transportable to unseen domains. 
In this work, we present a nonparametric strategy for learning invariant representations based on the recently-proposed Nadaraya-Watson (NW) head. 
The NW head makes a prediction by comparing the learned representations of the query to the elements of a support set that consists of labeled data.
% In the NW head, the prediction is made by comparing the learned representations of the query to the elements of a support set that consists of labeled data.
% In the NW head, the prediction is a weighted average of labels from a support set. 
% The weights are computed from distances between the query feature and support features.
We demonstrate that by manipulating the support set, one can encode different causal assumptions.
In particular, restricting the support set to a single environment encourages the model to learn invariant features that do not depend on the environment.
% That is, the conditional support set precludes the possibility of using spurious features to make a prediction.
We present a causally-motivated setup for our modeling and training strategy and validate on three challenging real-world domain generalization tasks in computer vision. 
\end{abstract}

\section{Introduction}
Machine learning models often fail when there is significant distribution shift. 
The goal of domain generalization is to be able to perform well with new distributions~\cite{kaddour2022causalsurvey,wang2022generalizing,zhou2022survey}.
In this work, we are interested in settings where multiple domains/environments are available during training and we have access to environment indicators.
A popular way to tackle domain generalization in this setting is to learn representations that are invariant across environments~\cite{jiang2022transportable,rojascarulla2018invariant,wang2022cisa}. 
The hope is that such representations will work well in, or are transportable to, unseen environments.
This invariance is often encoded via constraints on a learned predictor which aligns its behavior across environments; often, these conditions are derived using causal reasoning and/or by making assumptions about the data-generating process~\cite{pearl2009causality}.

In a parametric setting, almost all existing methods enforce these constraints by training a single model and adding a regularizer on top of a standard predictive loss~\cite{arjovsky2019irm,chevalley2022invariant,heinzedeml2019conditional,jiang2022transportable,krueger2021outofdistribution,sun2016coral,wald2021oodcalibration,wang2022desiderata}.
Most notably, invariant risk minimization (IRM) enforces the representations to be such that the optimal classifier on top of those representations is the same across all environments. 
Other examples include enforcing the layer activations of the predictor to be aligned across environments~\cite{sun2016coral}, enforcing the predictor to be calibrated across environments~\cite{wald2021oodcalibration}, and enforcing gradients of the predictor to be aligned across environments~\cite{shi2021fish}.
% In all aforementioned cases, the constraints attempt to align elements of predictors across environments.
Often, optimizing these constraints demand approximations or relaxations that undermine the efficacy of the approach~\cite{kamath2021does}.

In this work, we take a different approach using a nonparametric strategy based on a recently-proposed Nadaraya-Watson (NW) head~\cite{wang2023nwhead}.
Instead of computing the class probability directly from an input query, the NW head makes a prediction by comparing the learned representations of the query to the elements of a support set that consists of labeled data.
Thus, the NW prediction is computed \textit{relative to other real datapoints} in the support set, with the support set providing a degree of flexibility not possible with parametric models. 
In particular, one can manipulate it during training in a way which restricts the types of comparisons that the model can make.
% These manipulations can implicitly encode causal assumptions, and arguably represents a more natural mechanism to align predictors across environments.

In this work, we manipulate the support set during training to encode causal assumptions for the purposes of learning invariant representations.
Specifically, restricting the support set to be drawn from a single environment precludes the possibility of using environment-specific features to make a prediction for a given query.
We show that this setup is causally-motivated and relates to existing causal frameworks.
Furthermore, we show that this training strategy leads to competitive to superior results compared to state-of-the-art parametric baselines.

Our contributions are as follows:
\begin{itemize}
    \item We present causally-motivated assumptions for domain generalization which justify our modeling and training strategy.
    \item We present a novel approach to invariant representation learning using the nonparametric Nadaraya-Watson head, which can account for causal assumptions by manipulating a support set.
    In particular, we propose a training strategy which, unlike competing baselines, has \textit{no invariance hyperparameter to tune}.
    \item We validate our approach on several datasets and demonstrate competitive results compared to state-of-the-art parametric baselines.
\end{itemize}

\section{Related Works}
\subsection{Domain Generalization and Invariant Representations}
Domain generalization seeks to make models robust to unseen environments and is an active area of research~\cite{kaddour2022causalsurvey,wang2022generalizing,zhou2022survey}.
One line of work augments or synthetically-generates additional training images to increase robustness of learned features to unseen environments~\cite{wang2020heterogeneous,xu2019adversarial,yao2022lisa,yue2022domain,zhou2020generation}.
In particular, LISA uses a mixup-style~\cite{zhang2018mixup} interpolation technique to generate augmented images, which the authors demonstrate improves out-of-distribution robustness~\cite{yao2022lisa}.
Another line of work broadly seeks to align features across distributions.
Deep CORAL aligns correlations of layer activations in deep neural networks~\cite{sun2016coral}, and other works minimize the divergence of feature distributions with different distance metrics such as maximum mean discrepancy~\cite{tzeng2014deep,long2015learning}, an adversarial loss~\cite{ganin2016domainadversarial,li2018adversarial}, and Wasserstein distance~\cite{zhou2021optimaltransport}.
Still other works approach the problem from the perspective of the gradients and optimization~\cite{dou2019semantic,li2017learning,mansilla2021domain,shi2021fish,wang2023sharpnessaware}.
For example, Fish aligns the gradients from different domains~\cite{shi2021fish}.

One can also achieve domain generalization via learning invariant representations, which often requires reasoning about the data-generating process from a causal perspective to arrive at appropriate constraints~\cite{pearl2009causality}.
Invariant causal prediction (ICP) formulates the problem from a feature selection perspective, where the goal is to select the features which are direct causal parents of the label~\cite{peters2015causal}.
Invariant Risk Minimization (IRM) can be viewed as an extension of ICP designed for deep, nonlinear neural networks.
%As our work builds on IRM, we provide a more detailed description of it here.
The IRM objective can be summarized as finding the representation $\nwrep$ such that the optimal linear classifier's parameters $w^*$ on top of this representation is the same across all environments~\cite{arjovsky2019irm}.
% The constrained optimization problem:
% \begin{equation}
%     \min_{\rep, \hat{w}} \sum_{e \in E} \Rcal^e (\rep, \hat{w}) \ \ \ \text{s.t}. \ \ \ \hat{w} \in \argmin_w \Rcal^e (\rep, w), \ \forall e \in E
% \end{equation}
This bi-level program is highly non-convex and difficult to solve. 
To find an approximate solution, the authors consider a Lagrangian form, whereby the sub-optimality with respect to the constraint is expressed as the squared norm of the gradients of each of the inner optimization problems.
Follow-up works analyzing IRM have raised theoretical issues with this objective  and presented some practical concerns~\cite{gulrajani2020search,kamath2021does,rosenfeld2021risksofirm}.
Various flavors of IRM have also been proposed by introducing different regularization terms~\cite{krueger2021outofdistribution,wald2021oodcalibration, wang2022desiderata}.

% Other works have made various assumptions and proposed different regularization terms, including based on the variance of environment risks~\cite{krueger2021outofdistribution}, calibration across environments~\cite{wald2021oodcalibration}, alignment of conditional variances~\cite{heinzedeml2019conditional}, and capturing of necessary and sufficient causes~\cite{wang2022desiderata}.
% The relaxed IRM objective encodes the constraint in a Lagrangian form, but follow-up works have questioned the efficacy of IRM and its variants~\cite{kamath2021does}.

\subsection{Nonparametric Deep Learning}
Nonparametric models in deep learning have received much attention in previous work.
Deep Gaussian Processes~\cite{damianou2013gps}, Deep Kernel Learning~\cite{wilson2015kernellearning}, and Neural Processes~\cite{kossen2021npt} build upon Gaussian Processes and extend them to representation learning. 
Other works have generalized $k$-nearest neighbors~\cite{papernot2018deepknn,taesiri2022visual}, decision trees~\cite{zhang2018decisiontrees}, density estimation~\cite{fakoor2020trade}, and more general kernel-based methods~\cite{ghosh2001overviewrbfs,nguyen2021krr,zhang2021hybrid} to deep networks and have explored the interpretability that these frameworks provide.
Closely-related but orthogonal to nonparametric models are attention-based models, most notably self-attention mechanisms popularized in Transformer-based architectures in natural language processing~\cite{vaswani2017attention} and, more recently, computer vision~\cite{dosovitskiy2020vit,jaegle2021perceiver,parmar2018imagetransformer}.
% Retrieval models in NLP learn to look up relevant training instances for prediction~\cite{sachan2021retriever,borgeaud2021improving} in a nonparametric or semiparametric manner.
% Many works have recognized the inherent interpretable nature of attention and have leveraged it in vision~\cite{chen2018this,guan2018diagnose,wang2017residual,xu2015showattendtell}.
Nonparametric transformers apply attention in a nonparametric setting~\cite{kossen2021npt}.
% NPT leverages self- and cross-attention between both attributes and datapoints, which are stacked up successively in an alternating fashion to yield a deep, non-linear architecture.
%These operations are then stacked successively and alternatingly.

Recently, Wang et al. proposed the NW head~\cite{wang2023nwhead}, an extension of the classical NW model~\cite{bishop2006mlbook,nadaraya1964,watson1964} to deep learning.
In the NW head, the prediction is a weighted average of labels from a support set. 
The weights are computed from distances between the query and support features.
The NW head can yield better calibration and interpretability, with similar accuracy compared to the dominant approach of using a parametric classifier with fully-connected layers.
In this work, we leverage the NW head to encode causal assumptions via the support set.
The interpretability and explainability benefits of the NW head carry over in this work; while not of primary focus, we explore these properties in the Appendix.

\begin{figure}[t]
\centering
\includegraphics[width=\linewidth]{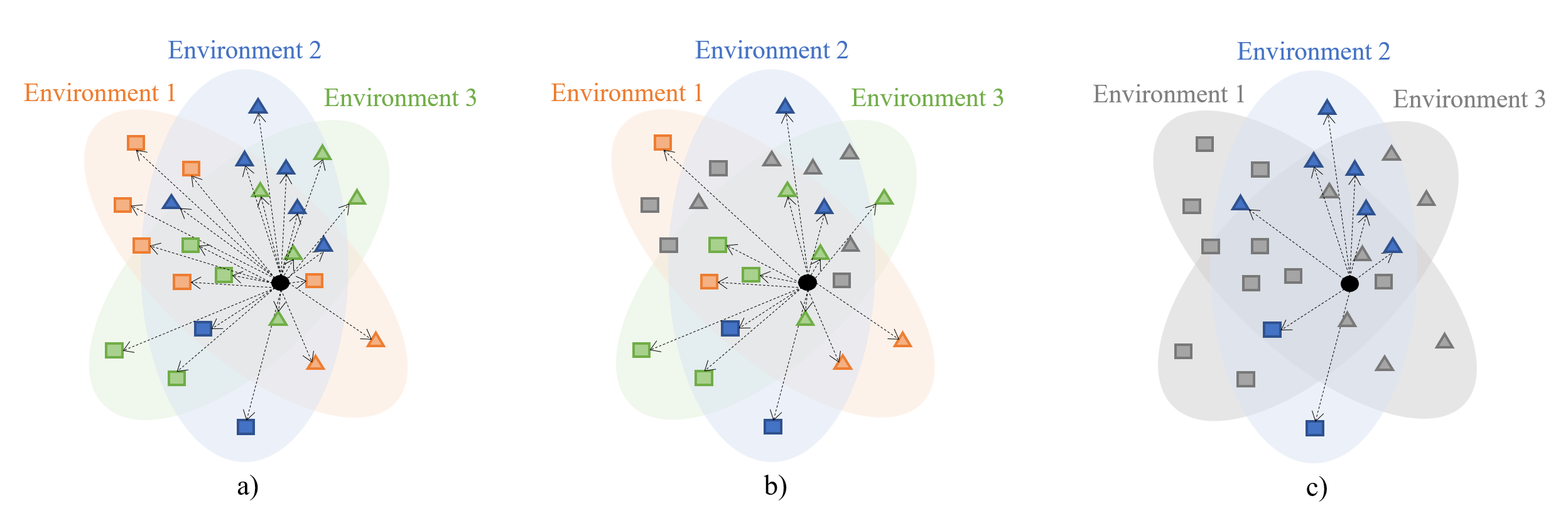}
\caption{
Illustration of proposed approach.
Support set of labeled datapoints (square/triangle) from 3 environments lie in 3 regions in the feature space.
Black circle denotes query datapoint with unknown label.
a) The NW head models $P(Y|X)$ by making predictions as a function of distances to labeled datapoints in the feature space (visualized as dotted arrows) . 
b) Balancing comparisons across labels for all environments models $P^B(Y|X)$. 
c) Conditioning on a single environment models $P_e(Y|X)$.}\label{fig:motivating}
\end{figure}

\section{Preliminaries}
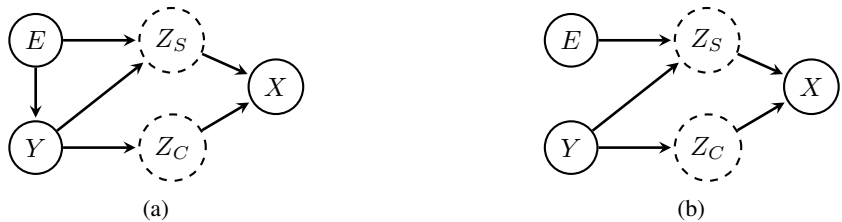
\begin{figure*}[h!]
    \centering
    \begin{subfigure}[t]{0.5\textwidth}
        \centering
        \begin{tikzpicture}[> = stealth, shorten > = 1pt, auto, node distance = 0.5cm, thick]
    \tikzstyle{every state}=[draw=black, thick, fill=white, minimum size=4mm]
    \node[state] (E) {$E$};
    \node[state] (Y) [below=0.7cm of E]{$Y$};
    \node[state] (Z_C) [right=1cm of Y,dashed] {$Z_C$};
    \node[state] (Z_S) [right=1cm of E,dashed] {$Z_S$};
    \node[state] (X) [right=1cm of $(Z_C.north)!0.25!(Z_S.north)$] {$X$};
    % \plate [inner sep=.3cm,xshift=.02cm,yshift=.2cm, dashed] {plate1} {(X_spurious) (X_invariant)} {$\phi(X)$}; % 
    \path[->, line width=1] (E) edge node {} (Y);
    \path[->, line width=1] (Y) edge node {} (Z_C);
    \path[->, line width=1] (Y) edge node {} (Z_S);
    \path[->, line width=1] (E) edge node {} (Z_S);
    \path[->, line width=1] (Z_C) edge node {} (X);
    \path[->, line width=1] (Z_S) edge node {} (X);
\end{tikzpicture}
        \caption{}
    \end{subfigure}%
    ~ 
    \begin{subfigure}[t]{0.5\textwidth}
        \centering
        \begin{tikzpicture}[> = stealth, shorten > = 1pt, auto, node distance = 0.5cm, thick]
    \tikzstyle{every state}=[draw=black, thick, fill=white, minimum size=4mm]
    \node[state] (E) {$E$};
    \node[state] (Y) [below=0.7cm of E]{$Y$};
    \node[state] (Z_C) [right=1cm of Y,dashed] {$Z_C$};
    \node[state] (Z_S) [right=1cm of E,dashed] {$Z_S$};
    \node[state] (X) [right=1cm of $(Z_C.north)!0.25!(Z_S.north)$] {$X$};
    % \plate [inner sep=.3cm,xshift=.02cm,yshift=.2cm, dashed] {plate1} {(X_spurious) (X_invariant)} {$\phi(X)$}; % 
    \path[->, line width=1] (Y) edge node {} (Z_C);
    \path[->, line width=1] (Y) edge node {} (Z_S);
    \path[->, line width=1] (E) edge node {} (Z_S);
    \path[->, line width=1] (Z_C) edge node {} (X);
    \path[->, line width=1] (Z_S) edge node {} (X);
\end{tikzpicture}
        \caption{}
    \end{subfigure}
%     \begin{subfigure}[t]{0.3\textwidth}
%         \centering
%         \begin{tikzpicture}[> = stealth, shorten > = 1pt, auto, node distance = 0.5cm, thick]
%     \tikzstyle{every state}=[draw=black, thick, fill=white, minimum size=4mm]
%     \node[state] (E) {$E$};
%     \node[state] (Y) [left=0.7cm of E]{$Y$};
%     \node[state] (Z_C) [below=1cm of Y,dashed] {$Z_C$};
%     \node[state] (Z_S) [below=1cm of E,dashed,fill=gray] {$Z_S$};
%     \node[state] (X) [below=1cm of Z_C] {$X$};
%     % \plate [inner sep=.3cm,xshift=.02cm,yshift=.2cm, dashed] {plate1} {(X_spurious) (X_invariant)} {$\phi(X)$}; % 
%     \path[->, line width=1] (Y) edge node {} (Z_C);
%     \path[->, line width=1] (Y) edge node {} (Z_S);
%     \path[->, line width=1] (E) edge node {} (Z_S);
%     \path[->, line width=1] (Z_C) edge node {} (X);
%     \path[->, line width=1] (Z_S) edge node {} (X);
% \end{tikzpicture}
%         \caption{}
    % \end{subfigure}
    \caption{
    a) Causal Directed Acyclic Graph (DAG) we consider in this work.
Solid nodes are observed and dashed nodes are unobserved.
% Shaded nodes are conditioned on.
We assume an anti-causal setting where label $Y$ causes $X$, and $X$ has 2 causal parents:
``style'' features, $Z_S$, which are influenced by the environment $E$; and environment-independent ``content'' features of $X$, $Z_C$, which are causally influenced by the label $Y$.
$E$ potentially influences $Y$.
Both $E$ and $Y$ have direct influence on style features $Z_S$.
b) Same DAG as a) with an intervention on $Y$.
We note that $Y \indep E \mid Z_C$ and $Y \not\indep E \mid Z_S$.
% c) Same DAG as b) with conditioning on $Z_S$. 
% Using d-separation, we see that $Y \indep E \mid C$ and $Y \not\indep E \mid S$.
    }
\label{fig:graphs}
\end{figure*}

\textbf{Problem Setting.}
Let $X, Y$ denote a datapoint and its corresponding discrete class, and $E$ denote the environment (or domain) where $X,Y$ originates.\footnote{We assume the support of $E$, $\supp(E)$, is finite.}
% Denote their sample spaces $\Xcal, \Ycal, \Ecal$, accordingly.
That is, the elements of the training dataset $\Dcal_{tr} = \{x_i, y_i, e_i\}_{i=1}^N$ are drawn first by sampling the discrete random variable $e_i \sim P(E)$, and then sampling $x_i, y_i \sim P(X, Y \mid E=e_i):=P_{e_i}(X, Y)$.
% from drawn training environments $E_{train}$, such that training data from each environment is i.i.d. sampled 
Our goal is to learn classifiers that will generalize to new, unseen environments.
% At test-time we are given $X_{test}$ and our goal is to predict $Y_{test}$, with the assumption that the environment $E_{test}$ from which they are drawn was not seen during training, i.e. $\supp(E) \cap \supp(E_{test}) = \emptyset$.
    
% We are concerned with anti-causal learning; that is, that $Y$ causes $X$.
% Let $\rep(X)$ denote the representation of $X$ in some feature space.
\textbf{Assumptions.}
We assume there exists a pair of latent causal parents of $X$: an environment-independent (``content'') factor $Z_C$ and an environment-dependent (``style'') factor $Z_S$.\footnote{This is sometimes referred to as style-content or causal decomposition~\cite{kaddour2022causalsurvey,rosenfeld2021risksofirm,wang2022cisa}.}
We assume the causal mechanism that generates $X$ from ($Z_C$, $Z_S$) is injective, so that, in principle, it is possible to recover the latent features from the observations; i.e. there exists a function $g$ such that $g(X) = (Z_C, Z_S)$.
We further assume that $g$ can be disentangled into $g_C$ and $g_S$, such that $(Z_C, Z_S) = g(X) = (g_C(X), g_S(X))$.
The causal graph is depicted in Fig.~\ref{fig:graphs}a.
Finally, we assume that if any $X=x$ has a non-zero probability in one environment, it has a non-zero probability in all environments.

%Finally, we assume that $X$ has the same support in all enviroments. In other words, if $X=x$ has a non-zero probability in one environment, it should have a non-zero prob in all environments.

\textbf{Motivation.} The motivating application in this work is image classification, where each realization of $E$ might represent a different site, imaging device, or geographical region where $X, Y$ are collected.
For example, in medical imaging, different hospitals ($E$) may collect images ($X$) attempting to capture the presence of some disease ($Y$), but may differ in their imaging protocols which lead to differences in the image style features $Z_S$ (e.g. staining, markings, orientation). 
In addition, we allow $Z_S$ to be influenced by the label itself (for example, positive examples are more likely to be marked by a doctor or have specific staining than negative examples).
Finally, the prevalence of $Y$ may be influenced by $E$ (for example, the prevalence of a disease may be higher in a certain hospital). 

The goal is to find an estimator for $Y$ which relies only on the direct causal links between $Y$ and $X$ and not on any spurious associations between $E$ and $X$, as these may change in a new, unseen environment.
That is, we seek an estimator which relies only on $Z_C$ and which is independent of $E$ or $Z_S$. 

First, we note the direct causal dependence $E \rightarrow Y$.
For example, a model can exploit this association by learning to over-predict majority classes in a certain environment.
One way to remove the direct dependence is by intervening on $Y$, thus removing incoming edges to $Y$.
This essentially corresponds to matching the environment-specific prior on $Y$ between environments, and results in the intervened graph in Fig.~\ref{fig:graphs}b.\footnote{
This may be interpreted as making the model robust to label shift, see~\cite{lipton2018detecting,schoelkopf2012causal}.
% Here, we are making the implicit assumption that that all possible test data that we deploy this model on follows the distribution in Fig.~\ref{fig:graphs}b. 
% While prior works have explicitly accounted for this~\cite{minh}, in the absence of extra information that enables one to exploit this correlation, we make the (reasonable) assumption . We discuss this point further in the Conclusion section.
}
% \footnote{In practice, this can be achieved by balancing the frequency of labels in the training dataset across environments.
% Note that this idea of data balancing, resampling, and reweighting is often performed in works without explicitly stating so, and is effective and previous works have brought it to light~\cite{wang2023causal}.}
Let us refer to any distribution which follows the DAG in Fig.~\ref{fig:graphs}b as $P^B_{e}(X, Y)$.

Second, we observe that there is a potential non-causal association flow between $E$ and $Y$ through the colliders $X$ and $Z_S$, when either one of these are conditioned on (i.e. are observed).
An estimator which relies on $Z_S$ potentially leaks information from $E$, and this is unstable in new environments.
Reading d-separation on this intervened graph, we infer that $Y \indep E \mid Z_C=g_C(X)$ and $Y \not\indep E \mid Z_S=g_S(X)$, that is:
% We would like to learn a representation $\rep: \Xcal \rightarrow \Phi$ which captures $Z_C$    
\begin{equation}
P^B_{e}(Y\mid g_C(X)) = P^B_{e'}(Y\mid g_C(X)) \ \ \forall e, e' \in E.
\label{eq:suff_inv}
\end{equation}
In words, this assumption states that the probability of $Y$ given the environment-invariant parts of $X$ is the same across any environment $e \in E$. 
% \footnote{This is sometimes referred to as the sufficiency invariance in the literature~\cite{wang2022cisa,wald2021oodcalibration}.}

Thus, we seek an estimator that 1) enables interventions on $Y$ such that the direct dependence $E \rightarrow Y$ can be removed, and 2) can further encode the assumption in Eq.~\eqref{eq:suff_inv}.
% We propose an elegant method to estimate $P(Y\mid C, E)$ and enforce the invariance in Eq.~\eqref{eq:suff_inv}, which we detail next.

\section{NW Head for Invariant Prediction}

\begin{figure}[t]
\centering
\includegraphics[width=\linewidth]{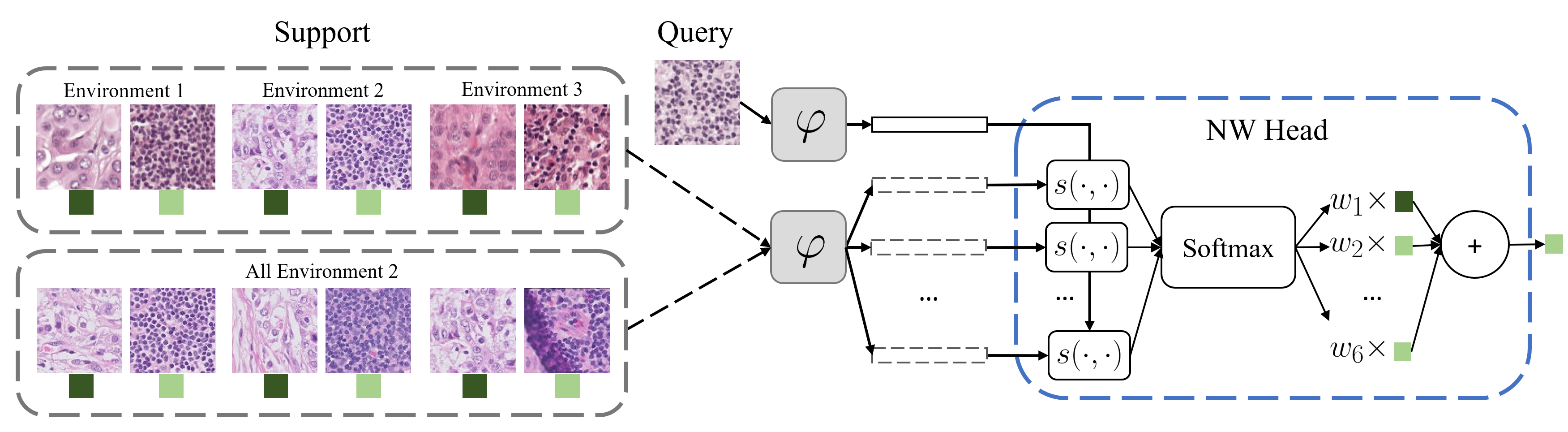}
\caption{
A depiction of the NW head on a tumor detection task. 
The NW head computes Euclidean distances $s(\cdot, \cdot)$ between query and support features, and uses the distances to weight the support labels. Colored squares represent labels.
Diagram displays two different support sets. 
Top is unconditional support, where support data is drawn from the training data without knowledge of environment information. 
Bottom is an example of a manipulated support where all support data is drawn from a fixed environment (note similarity in color). 
Such a support set precludes the possibility of using environment-specific features to make a prediction.
}\label{fig:arch}
\end{figure}

% Suppose we have a dataset of images, corresponding one-hot labels, and (discrete) attributes of interest: $\mathcal{D}^e = \{(x_i^e, \vec{y}_i^e\}_{i=1}^N$.
Given a datapoint $x$, support set $\Scal = \{x_i, y_i\}_{i=1}^{N_s}$, and parameters $\rep$, the NW head estimates $P(Y=y\mid X=x)$ by outputting a prediction formulated as a weighted sum of support set labels, where the weights are some notion of similarity in the feature space~\cite{wang2023nwhead}:
\begin{equation}
\label{eq:nwhead}
    \hat{P}(Y=y\mid X=x; \rep) := \NW_\rep(x, \Scal) =
    \frac{\sum_{i=1}^{N_s} \exp{\{ s(\rep(x), \rep(x_i)) \}\vec{y_i}}}{\sum_{j=1}^{N_s} \exp{\{s(\rep(x), \rep(x_j))\}}}.
\end{equation}
Here, $\vec{y}$ is the one-hot encoded version of $y$ and $s(\cdot, \cdot)$ is a similarity/kernel function that captures the similarity between pairs of features.
In this work, we set $s$ as the negative Euclidean distance.
A graphical depiction is shown in Fig.~\ref{fig:arch}.
% The kernel may be fixed or have its own learnable parameters $\phi$.

Manipulating the support set can encode certain causal assumptions.
We consider two types of manipulations:
\begin{enumerate}
    \item Balancing classes in $\Scal$, denoted $\Scal^B$ (see Fig.~\ref{fig:motivating}b). This can be interpreted as an intervention on $Y$, and removes the dependence on $E \rightarrow Y$, i.e.:
    \begin{equation}
    \label{eq:nwb}
        \hat{P}^B(Y=y \mid X=x; \rep) := \NW_\rep(x, \Scal^B    ).
    \end{equation}
    \item Conditioning $\Scal$ on a single environment, denoted $\Scal_e$ (see Fig.~\ref{fig:motivating}c). This can be interpreted as conditioning the probability estimate on $E=e$, i.e.:
\begin{equation}
    \label{eq:nwe}
    \hat{P}_e(Y=y \mid X=x; \rep) := \NW_\rep(x, \Scal_e).
\end{equation}
% That is, $E$ is fed as input \textit{implicitly via the support set}.
\end{enumerate}
Note that both balancing and conditioning can be achieved simultaneously, which we denote $\Scal^B_e$.
% We refer to any NW variant which encodes causal assumptions via manipulations of the support set as the causal NW head (\modelname).

% In addition, we propose a conditional variant of the NW head (abbreviated CNW) which estimates $P(Y|\rep_\theta(X), E)$ by conditioning the support set $\{X_i, Y_i\}_{i=1}^N$ on $E$:
% \begin{equation}
    % \hat{P}(Y \mid \rep_\theta(X), E) = f_\theta(x, \{X_i, Y_i \mid E\}_{i=1}^N).
% \end{equation}
% Then, $f_e$ is an estimator of $P(Y|X, E)$

\subsection{Objective and Optimization Details} 
Given a dataset of samples $\Dcal_{tr} = \{x_i, y_i, e_i\}_{i=1}^N$, we wish to leverage the NW head as a conditional estimator for $Y$ conditioned on $Z_C = g_C(X)$, where $g_C(X)$ is characterized by Eq.~\ref{eq:suff_inv}. 
This necessitates an optimization over both $\rep$ and the space of functions $g_C$.
Thus, we solve the following constrained maximum likelihood over $\rep$ and $g_C$:
% \begin{gather}
%    \max_{\rep, g_C}  \sum_{i=1}^N \log \hat{P}^B_{e_i} (Y_i \mid g_C(X_i); \rep) \\
% \text{s.t.} \ 
%     \hat{P}^B_e (Y_i \mid g_C(X_i)) = \hat{P}^B_{e'} (Y_i \mid g_C(X_i)) \ \ \forall i \in [N], \ \forall e, e'\in E.
% \end{gather}
\begin{align}
\begin{split}
\label{eq:objective}    
   &\argmax_{\rep, g_C}  \sum_{i=1}^N \log \hat{P}^B_{e_i} (y_i \mid g_C(x_i); \rep) \\
&\text{s.t.} \ 
    \hat{P}^B_e (y_i \mid g_C(x_i); \rep) = \hat{P}^B_{e'} (y_i \mid g_C(x_i); \rep), \ \ \forall i \in \{1, ..., N\}, \ \forall e, e'\in E.
\end{split}
\end{align}
Note that Eq.~\ref{eq:suff_inv} implies that $P^B_e (y_i \mid g_C(x_i)) = P^B (y_i \mid g_C(x_i))$.
Thus, the objective is equivalent to unconstrained maximum likelihood under the assumption in Eq.~\ref{eq:suff_inv}.

Instead of solving for $g_C$ explicitly, we let both $\rep$ and $g_C$ be related by the composition $\nwrep = \rep \circ g_C$, and set $\nwrep$ to be the learnable mapping of the NW head, i.e. a neural network.
Then, the objective becomes:
\begin{align}
\begin{split}
   &\argmin_{\nwrep}  \sum_{i=1}^N L(\NW_\nwrep(x_i, \Scal_{e_i}^B), y_i) \\
&\text{s.t.} \ 
    f_\nwrep(x_i, \Scal_e^B) = f_\nwrep(x_i, \Scal_{e'}^B), \ \ \forall i \in \{1, ..., N\}, \ \forall e, e'\in E,
\end{split}
\end{align}
where $L$ is the cross-entropy loss.
To make the objective tractable, we consider two possible variants:
\begin{enumerate}
\item
\textbf{Explicit.}
Solve the optimization problem explicitly via a Lagrangian formulation:
\begin{align}
\label{eq:nw_explicit}
   \argmin_{\nwrep} \sum_{i=1}^N L(\NW_{\nwrep} (x_i, \Scal^B_{e_i}), y_i) +
    \lambda \sum_{e, e'\in E} \sum_{i=1}^N \| \NW_\nwrep(x_i, \Scal_e^B) - \NW_\nwrep(x_i, \Scal_{e'}^B) \|_2^2.
\end{align}
where $\lambda > 0$ is a hyperparameter.

 \item
\textbf{Implicit.} 
Relax the optimization problem into the following unconstrained problem:
\begin{align}
\label{eq:nw_implicit}
   \argmin_{\nwrep} \sum_{e\in E} \sum_{i=1}^N L(\NW_\nwrep(x_i, \Scal_e^B), y_i).
   % = \max_{\rep, g_C} \sum_{e\in E} \sum_{i=1}^N \log \NW(\rep_\theta(X_i), \Scal^B_e)
\end{align}
In this formulation, the constraint will be approximately satisfied in the sense that the model will be encouraged to predict the ground truth for a given image, which is identical across all environments.
In practice, how well the solution satisfies the constraint will depend on model capacity, the data sample, and optimization procedure.

\end{enumerate}

\subsection{Optimization Details} 

During training, the support set $\Scal$ is drawn stochastically from the training set $\Dcal_{tr}$, and all queries and support datapoints are passed through the feature extractor $\nwrep$.
For computational efficiency, instead of sampling a unique support mini-batch at the query-level, we sample a unique support at the mini-batch level.
Thus, if $N_q$ and $N_s$ are the query and support mini-batch sizes respectively, the effective mini-batch size is $N_q + N_s$, instead of $N_q N_s$.
For the implicit variant, we sample one support set for a given mini-batch of queries, forward pass through the NW head, and compute the loss in Eq.~\eqref{eq:nw_implicit}.
For the explicit variant, we sample two support sets for a given mini-batch of queries, perform two independent forward passes through the NW head for each support set, and compute the loss in Eq.~\eqref{eq:nw_explicit}.
As discussed in prior work~\cite{wang2023nwhead}, the support batch size is a hyperparameter analogous and orthogonal to the query batch size.
% In practice, we choose both hyperparameters to maximize the GPU memory.
% To make the objective amenable to stochastic gradient descent, we sample a environment-specific support mini-batch in addition to a query mini-batch.

A technical point is that 
%each class in the query must have a matching class in the support; that is, 
the set of labels in the support mini-batch must cover the set of labels in the query mini-batch.
Thus, in our implementation, for $\Scal^B$, we cycle through all classes and randomly draw $N_c$ examples per class to include in the support.
For tasks with a large number of classes, one can subsample from the total number of classes, so long as the sampled classes cover the set of query classes.

\subsection{Inference modes}
Similar to how the support set can be manipulated during training, we can also design different inference strategies corresponding to different configurations of the support set at test-time.
We explore several different inference modes which are possible under the NW framework:
\begin{enumerate}
\item 
\textbf{Random.} Sample uniformly at random over the dataset, such that each class is represented $k$ times. 
% $\mathcal{S} \sim \mathcal{D}_{tr}$ and $|\mathcal{S}| = k|\mathcal{C}|$.

\item 
\textbf{Full.} 
Use the entire balanced training set.
% : $\mathcal{S}=\mathcal{D}_{tr}$.

\item
\textbf{Ensemble.}
Given the set of balanced features computed from Full mode, partition the computed features for all training datapoints by environment, compute the softmax predictions with respect to each environment balanced across labels, and average the predictions.

\item
\textbf{Cluster.}
Given the set of balanced features computed from Full mode, perform $k$-means clustering on the features of the training datapoints for each class.
These $k$ cluster centroids are then used as the support features for each class.  
This can be viewed as a distillation of the full training set for efficient inference, with the caveat that the support set no longer corresponds to observed datapoints.

\end{enumerate}
While Full, Ensemble, and Cluster require computing features for the entire support set, in practice these features and centroids can be precomputed.
In our experiments, we find that Cluster mode can be a sufficient replacement to Full mode, while being computationally cheaper. 
These inference modes can be used interchangeably and flexibly. 
As an example, consider a workflow which would involve using Cluster mode to perform efficient inference, and then using Full mode on a select few (potentially problematic) test queries to understand model behavior. 
% We stress that this flexibility at inference time is not possible with FC baseline models. 

\subsection{Connections to Prior Work}
Our assumptions in Eq.~\eqref{eq:suff_inv} are common across many works related to learning invariant predictors~\cite{peters2015causal,krueger2021outofdistribution,wald2021oodcalibration,rojascarulla2018invariant,koyama2021invariance}. 
Representatively, under the binary classification setting, the IRM objective finds a representation function $\nwrep$ which elicits an invariant predictor across environments $E$ such that for all $h$ that has a non-zero probability for $\nwrep(X)$ in any (and all) environment(s):
$$\E_e[Y \mid \nwrep(X)=h] = \E_{e'}[Y \mid \nwrep(X)=h], \ \forall e, e' \in E.$$
Eq.~\eqref{eq:suff_inv} can be viewed as a generalization of this equality to multi-class settings.%
\footnote{This equality has also been called "sufficiency invariance"~\cite{wang2022cisa}.}.

Furthermore, note that given the feature extractor $\nwrep$, the NW mechanism $f$ is a nonlearnable classifier, whereas $w$ is learned in the IRM setting. 
Thus, our proposed objective can be interpreted as learning invariant features $\nwrep$, where the \textit{fixed classifier constraint is satisfied by construction}. 
This avoids the need to approximate the complex bilevel optimization problem with a regularizer which assumes convexity and requires computing the Hessian.
Essentially, $f$ enforces invariance through the manipulation of the support set, providing a more intuitive and computationally simpler objective to optimize.

In the Experiments section, we compare IRM against a variant of our algorithm where we freeze the learned representations and finetune a linear classifier on top using the same training data.
We find that our algorithm performs better than IRM on all datasets, suggesting that it captures invariant representations better than IRM.

\section{Experiments and Results}

\input{tables/table_datasets}

\input{tables/table_results}

\subsection{Baselines}
We compare against several popular and competitive baseline algorithms:
empirical risk minimization (ERM)~\cite{vapnik1999erm}, invariant risk minimization (IRM)~\cite{arjovsky2019irm}, deep CORAL~\cite{sun2016coral}, Fish~\cite{shi2021fish}, LISA~\cite{yao2022lisa}, and CLOvE~\cite{wald2021oodcalibration}.
When available, results on baselines are pulled from their respective papers.
Details on baseline algorithms are provided in the Appendix.

\subsection{Datasets}
We experiment on 3 real-world domain generalization tasks.
Two are from the WILDS benchmark~\cite{koh2021wilds}, and the third is a challenging melanoma detection task.
Details on the datasets are summarized in Table~\ref{tab:datasets}, and further information is provided in the Appendix.

\begin{enumerate}
\item The Camelyon-17 dataset~\cite{bandi2019camelyon} comprises of microscopic images of stained tissue patches from different hospitals, where the label corresponds to presence of tumor tissue in patches and the environment is the hospital where the patch comes from.
\item The melanoma dataset is from the International Skin Imaging Collaboration (ISIC) archive\footnote{https://www.isic-archive.com}.
The ISIC dataset comprises of dermoscopic images of skin lesions from different hospitals, where the label corresponds to whether or not the lesion is diagnosed as melanoma and the environment is the hospital where the image comes from.
There is significant less positive examples and negative examples, with this label imbalance varying across environments (see Appendix). 
\item The Functional Map of the World (FMoW) dataset~\cite{christie2018functional} comprises of RGB satellite images, where the label is one of 62 building or land use categories, and the environment represents the year the image was taken and its geographical region.

\end{enumerate}
\subsection{Experimental Setup}
For each model variant, we train 5 separate models with different random seeds, and perform model selection on an out-of-distribution (OOD) validation set. 
For WILDS datasets, we follow all hyperparameters, use model selection techniques, and report metrics as specified by the benchmark.
This includes using a DenseNet-121 backbone initialized with pretrained ImageNet weights as $\nwrep$ and no random augmentations for both datasets.
Similarly for ISIC, we use a pretrained ResNet-50 backbone as $\nwrep$ with no augmentations, and perform model selection on an OOD validation set. 
Due to significant label imbalance, we report F1-score instead of average accuracy.

For NW algorithms, we refer to models which balance classes (i.e. modeling Eq.~\eqref{eq:nwb}) as \nwb{}, and models which additionally condition on environment (i.e. modeling both Eq.~\eqref{eq:nwb} and Eq.~\eqref{eq:nwe}) as \nwbe{}. 
For \nwbe{} models, we train explicit and implicit variants.
For all NW algorithms, we perform evaluation on all inference modes.
In addition, for completeness, we experiment on a variant where we freeze the feature extractor and finetune a linear probe on top of the learned representations on the same training data $\Dcal_{tr}$, which we refer to as ``Probe''.
As an example, the implicit variant of \nwbe{} is trained on Eq.~\eqref{eq:nw_implicit}, where the support set is balanced across classes (B) and conditioned on an environment (e).

We set $N_c=8$ for Camelyon-17 and ISIC and $N_c=1$ for FMoW. 
An analysis of this hyperparameter is provided in the Appendix.
The query batch size $N_q$ is set to 8 for all NW experiments. 
For Random and Cluster inference modes, we set $k=3$.
This was chosen based on prior work~\cite{wang2023nwhead}, where $k=3$ was shown to balance good error rate performance with computational efficiency.
For explicit variants, we tune $\lambda$ for all datasets via grid search on a held-out validation set.
Full hyperparameter details are provided in the Appendix.

All training and inference is done on an Nvidia A6000 GPU and all code is written in Pytorch.\footnote{Our code is available at \url{https://github.com/alanqrwang/nwhead.}}.

\subsection{Results}
Table~\ref{tab:results} shows the main results.
We find that on Camelyon-17 and ISIC datasets, \nwbe{} with Full mode outperforms all baselines and variants we consider.
In addition, \nwbe{} variants typically have lower variance across random seeds as compared to baselines.
For FMoW, \nwbe{} with Ensemble mode performs around $2\%$ lower than the best performing baseline, CLoVE.
We observe that the most computationally-efficient inference mode, Cluster, performs comparably to Full mode for \nwbe{} models, and is in fact the highest-performing model for ISIC.
Thus, we conclude that the support set can be an efficient replacement for Full.

For ISIC, we find that almost all \nwb{} modes (except Random) perform $\sim3\%$ better than ERM.
This may be attributed to balancing classes across environments, which we suspect has added benefit for highly imbalanced tasks.
In contrast, this boost is less apparent for Camelyon-17, which has relatively balanced classes.
As an ablation, we compare \nwb{} against an NW variant without class-balancing in the Appendix. 
\nwbe{} further improves over \nwb{} by $\sim7\%$.
Exploring further, we compare \nwb{} against an ERM variant with balanced classes per environment, which we denote \ermb{}.
This achieves $63.0\pm2.5$, which is on-par with 
\nwb{}. This is expected as the theoretical assumptions are the same for both models.

Comparing implicit to explicit variants of \nwbe{}, we do not find much difference in explicitly enforcing Eq.~\ref{eq:suff_inv}, although we do observe significantly lower variances across model runs.
Generally, we find the slight performance gain of explicit training to not be worth the computational overhead of doubling the number of support set forward passes per gradient step and tuning the hyperparameter $\lambda$.

While not the highest-performing, we highlight the consistent $1$-$5\%$ improvement of Probe models over IRM, indicating that \nwbe{} may be better at capturing invariant features.
However, other non-parametric inference modes still outperform Probe, possibly indicating that the learned features are more suitable for NW-style classifiers.

\section{Discussion and Limitations}
There are several advantages of the proposed NW approach over previous works.
First, the implicit training strategy in Eq.~\eqref{eq:nw_implicit} has no hyperparameter to tune, 
while remaining competitive with and often outperforming state-of-the-art baselines which all require tuning a hyperparameter coefficient in the regularized loss.
Second, the NW head enables interpretability by interrogating nearest neighbors in the feature space. Since these neighbors directly contribute to the model’s prediction (Eq.~\eqref{eq:nwhead}), interrogation enables a user to see what is driving the model’s decision-making. This not only allows for greater model transparency, but also enables interrogating the quality of the invariant features. 
We explore this capability in Section~\ref{sec:interpretability} in the Appendix. 
Note this this degree of transparency is not present in parametric baselines.
Lastly, from an intuitive standpoint, we believe our non-parametric approach to enforcing invariance across environments is more natural than baseline methods, since an environment is encoded by manipulating the support set to contain real samples only from that environment.
Other baseline methods resort to proxy methods to enforce invariance~\cite{wald2021oodcalibration,sun2016coral,krueger2021outofdistribution}.

One important limitation of our method is computational (see Appendix for analysis of runtimes).
The proposed approach requires pairwise comparisons, which scales quadratically with sample size.
Practically, this means passing a separate support mini-batch in addition to a query mini-batch at every training iteration.
This limitation is compounded for explicit variants, in which two sets of support sets must be drawn independently.
Future work may explore more computationally-efficient training strategies. 
At inference time, Full, Cluster, and Ensemble modes are expensive procedures which require computing features for the entire support set, although precomputing features can mitigate this.
However, we argue that in high-risk, safety-critical domains like medical imaging, high inference throughput may not be as important as performance, interpretability, and robustness.

We expect the proposed approach to work well with tasks that have several (and diverse) sets of examples per label class in each environment. 
If this is not the case, as in the FMoW dataset, the resulting model will be sub-optimal.
In particular, in the extreme case where no example is present for a specific class in a given environment, constructing a support set with labels that cover the ground truth label of the query images will not always be possible.
This will, in turn, impact performance.

% We find that our model excels at tasks with a relatively small number of distinct classes and where the image diversity of the task is relatively low, which has been observed in past work~\cite{wang2023nwhead}.
% This scenario is common in medical imaging tasks, where inter- and intra-class variability is relatively low.
% Potentially, this strengthens the presence of the edge $Y \rightarrow Z_S$ in our causal DAG.
% Future work may explore improving performance for high-diversity tasks, possibly through different causal assumptions or different kernels.

\section{Conclusion}
We presented a nonparametric strategy for invariant representation learning based on the Nadaraya-Watson (NW) head. 
In the NW head, the prediction is made by comparing the learned representations of the query to the elements of a support set that consists of labeled data.
% In the NW head, the prediction is a weighted average of labels from a support set. 
% The weights are computed from distances between the query feature and support features.
We demonstrated two possible ways of manipulating the support set, and demonstrate how this corresponds to encoding different assumptions from a causal perspective.
% For example, restricting the support set to a single environment encourages the model to learn features independent of environment (e.g. color, lighting, orientation, etc.).
% That is, the conditional support set precludes the possibility of using spurious features to make a prediction.
We validated our approach on three challenging and real-world datasets. 

We believe there are many interesting directions of further research.
First, our treatment is restricted to classification tasks. 
Future work may explore an extension to the regression setting.
Second, it can be interesting to explore adaptation to the test domain, given additional information. 
For example, reweighting the occurrence of samples per label could provide improved results given knowledge about the edge $E \rightarrow Y$ in the test distribution.
One can further envision implementing the proposed method in settings where there are previously unseen test time labels/tasks.
Finally, we are interested in replacing the fixed similarity function with a learnable kernel. 

\section*{Acknowledgements}
Funding for this project was in part provided by the NIH grant R01AG053949, and the NSF CAREER 1748377 grant.

%Third, an NW-style prediction head, unlike a fully-connected head, can in theory be trained for 
%Incorporating labels which are unshared across all training environments could lead to more general and robust features.
% Previous works have explored an NW prediction head for few-shot and zero-shot classification tasks, and a 
%Lastly, future work might explore open-set recognition, where there may be unseen labels 

% \section*{References}

\bibliographystyle{plainnat}
\bibliography{bib.bib}

% References follow the acknowledgments in the camera-ready paper. Use unnumbered first-level heading for
% the references. Any choice of citation style is acceptable as long as you are
% consistent. It is permissible to reduce the font size to \verb+small+ (9 point)
% when listing the references.
% Note that the Reference section does not count towards the page limit.
% \medskip

% {
% \small

% [1] Alexander, J.A.\ \& Mozer, M.C.\ (1995) Template-based algorithms for
% connectionist rule extraction. In G.\ Tesauro, D.S.\ Touretzky and T.K.\ Leen
% (eds.), {\it Advances in Neural Information Processing Systems 7},
% pp.\ 609--616. Cambridge, MA: MIT Press.

% [2] Bower, J.M.\ \& Beeman, D.\ (1995) {\it The Book of GENESIS: Exploring
%   Realistic Neural Models with the GEneral NEural SImulation System.}  New York:
% TELOS/Springer--Verlag.

% [3] Hasselmo, M.E., Schnell, E.\ \& Barkai, E.\ (1995) Dynamics of learning and
% recall at excitatory recurrent synapses and cholinergic modulation in rat
% hippocampal region CA3. {\it Journal of Neuroscience} {\bf 15}(7):5249-5262.
% }
%%%%%%%%%%%%%%%%%%%%%%%%%%%%%%%%%%%%%%%%%%%%%%%%%%%%%%%%%%%%

\clearpage
\appendix
\section{Description of Baselines}
\begin{itemize}
    \item Empirical risk minimization (ERM)~\cite{vapnik1999erm} minimizes the sum of errors across domains and examples.
    \item Invariant risk minimization (IRM)~\cite{arjovsky2019irm} learns a feature representation such that the optimal linear classifier on top of that representation matches across domains. 
    For WILDS datasets, we pull baseline performance from~\cite{wald2021oodcalibration}. 
    For ISIC, we use the implementation from ~\cite{gulrajani2020search}.
    \item Deep CORAL~\cite{sun2016coral} penalizes differences in the means and covariances of the feature distributions (i.e., the distribution of last layer activations in a neural network) for each domain. 
    For WILDS datasets, we pull baseline performance from~\cite{wald2021oodcalibration}. 
    For ISIC, we use the implementation from ~\cite{gulrajani2020search}.
    % Conceptually, CORAL is similar to other methods that encourage feature representations to have the same distribution across domains (Tzeng et al., 2014; Long et al., 2015; Ganin et al., 2016; Li et al., 2018c,b).
    \item Fish targets domain generalization by maximizing the inner product between gradients from different domains.
    For WILDS datasets, we pull baseline performance from the original paper. 
    For ISIC, we use the implementation from ~\cite{gulrajani2020search}.
    \item LISA augments the set of training data by randomly performing two types of mixup-style~\cite{zhang2018mixup} interpolations: intra-label (same label, different domain) and inter-label (same domain, different label).
    For WILDS datasets, we pull baseline performance from the original paper, and train their implementation on the ISIC dataset.
    \item CLOvE~\cite{wald2021oodcalibration} finds an invariant classifier by enforcing the classifier to be calibrated across all training domains.
    While the original paper proposes several model variants leveraging this idea, we report their best-performing variant, which starts with a trained CORAL model and finetunes the weights using a regularized cross-entropy loss. 
    The regularizer aggregates Maximum Mean Calibration Error (MMCE)~\cite{kumar2018mmce} over all training domains.
    For WILDS datasets, we pull baseline performance from~\cite{wald2021oodcalibration}. 
    As their implementation is not publicly-available, we implement it for ISIC.
\end{itemize}

\section{Description of Datasets}
\begin{figure}[h]
\centering
\subcaptionbox{Camelyon-17\label{cw_10}}{%
  \includegraphics[width=\textwidth]{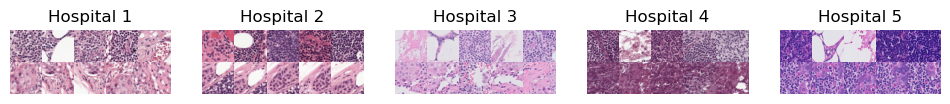}%
  }\par\medskip
\subcaptionbox{ISIC\label{cw_25}}{%
  \includegraphics[width=\textwidth]{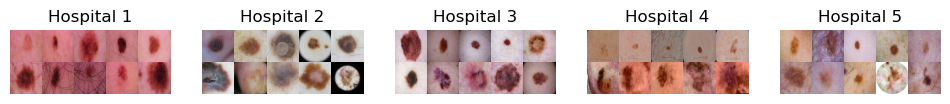}%
  }\par\medskip        
\subcaptionbox{FMoW\label{cw_50}}{%
  \includegraphics[width=\textwidth]{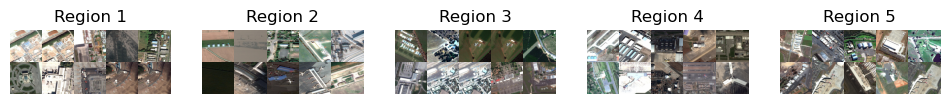}%
  }
\caption{Representative images for datasets, separated by domain. Each row depicts a separate class. For FMoW, for simplicity, we show 2 classes out of 62 and only images before 2013.}
\label{fig:images}
\end{figure}

Representative examples of the 3 datasets are shown in Fig.~\ref{fig:images}.
\begin{itemize}
\item \textbf{Camelyon-17.}
We use Camelyon-17 from the WILDS benchmark~\cite{bandi2019camelyon,koh2021wilds}, which provides 450,000 lymph-node scans sampled from 5 hospitals. 
Camelyon-17 is a medical image classification task where the input $x$ is a 96 × 96 image and the label $y$ is whether there exists tumor tissue in the image. 
The environment denotes the hospital that the patch was taken from. 
The training dataset is drawn from the first 3 hospitals, while out-of-distribution validation and
out-of-distribution test datasets are sampled from the 4th hospital and 5th hospital, respectively.
% the input x is a 96x96 histopathological image, the label y
% is a binary indicator of whether the central 32x32 region contains any tumor tissue, and the domain
% d is an integer that identifies the hospital that the patch was taken from.
% Data. The dataset comprises 450,000 patches extracted from 50 whole-slide images (WSIs) of
% breast cancer metastases in lymph node sections, with 10 WSIs from each of 5 hospitals in the
% Netherlands. Each WSI was manually annotated with tumor regions by pathologists, and the resulting
% segmentation masks were used to determine the labels for each patch. We also provide metadata
% on which slide (WSI) each patch was taken from, though our baseline algorithms do not use this
% metadata.
% We split the dataset by domain (i.e., which hospital the patches were taken from):
% 1. Training: 302,436 patches taken from 30 WSIs, with 10 WSIs from each of the 3 hospitals in the
% training set.
% 2. Validation (OOD): 34,904 patches taken from 10 WSIs from the 4th hospital. These WSIs are
% distinct from those in the other splits.
% 3. Test (OOD): 85,054 patches taken from 10 WSIs from the 5th hospital, which was chosen because
% its patches were the most visually distinctive. These WSIs are also distinct from those in the
% other splits.
% 4. Validation (ID): 33,560 patches taken from the same 30 WSIs from the training hospitals.

\item \textbf{ISIC.}
The melanoma dataset is from the International Skin Imaging Collaboration (ISIC) archive\footnote{https://www.isic-archive.com}.
Data from the archive are collected by different organizations %(Memorial Sloan Kettering Cancer Center, Medical University of Vienna, Hospital Clínic de Barcelona, Melanoma Institute Australia, the University of Queensland, Boston University)
at different points in time~\cite{codella2019skin,codella2018skin,combalia2019bcn20000,gutman2016skin,rotemberg2021patient,scope2009dermoscopic,tschandl2018ham10000}.
There are about 70k data samples in total.
In particular, the resized input image $x$ is a 224 × 224 image and a binary target label $y$ denotes whether the image exhibit is melanoma or not. 
The environment is the hospital from which the image was collected~\footnote{Hospitals are Hospital Clinic of Barcelona, Medical University of Vienna, University of Queensland Diamantina Institute, Memorial Sloan Kettering Cancer Center, University of Sydney Melanoma Diagnostic Centre, and University of Pittsburgh Medical Center.}.
% We consider Site (there are 8 different sites, listed in Appendix~\ref{app:data_example}) as $Z$,
% which may be spuriously correlated with $Y$~\cite{lian2017natural} (Figure~\ref{fig:isic_setups}).
%The second is \textit{Age} which is arguably a possible cause of $Y$~\cite{paulson2020age} (Figure~\ref{fig:isic_setups}, right panel).
%The values of \textit{Age} in ISIC are discretized so \textit{Age} is a categorical variable.
% Samples are grouped into environments based on spatio-temporal information as shown in Table~\ref{tab:isic_site}.
% Table~\ref{tab:isic_site} also shows that the marginal prevalence of melanoma, $P(Y=1|E)$, varies drastically between environments.
We follow a similar setup to Camelyon-17.
The training dataset is drawn from the first 3 hospitals, while out-of-distribution validation and out-of-distribution test datasets are sampled from the 4th hospital and 5th hospital, respectively.
For preprocessing, we filter out datapoints that are not specifically categorized as "benign" or "malignant" (e.g. "indeterminate").
The OOD validation dataset is from the "Barcelona1" site indicator and the OOD test dataset is the "Vienna1" site indicator.
% We resize all images to 224 × 224.

\item \textbf{FMoW.}
The FMoW dataset is from the WILDS benchmark~\cite{christie2018functional,koh2021wilds}, a satellite image classification task which includes 62 classes and 80 domains (16 years x 5 regions). 
Concretely, the input $x$ is a 224 × 224 RGB satellite image, the label $y$ is one of the 62 building or land use categories, and the environment represents the year that the image was taken as well as its corresponding geographical region – Africa, the Americas, Oceania, Asia, or Europe. 
The train/test/validation splits are based on the time when the images are taken. 
Specifically, images taken before 2013 are used as the training set. 
Images taken between 2013 and 2015 are used as the validation set. Images taken after 2015 are used for testing.
\end{itemize}

\clearpage
\section{Hyperparameter Details}
\begin{figure}[h]
\centering
\includegraphics[width=\linewidth]{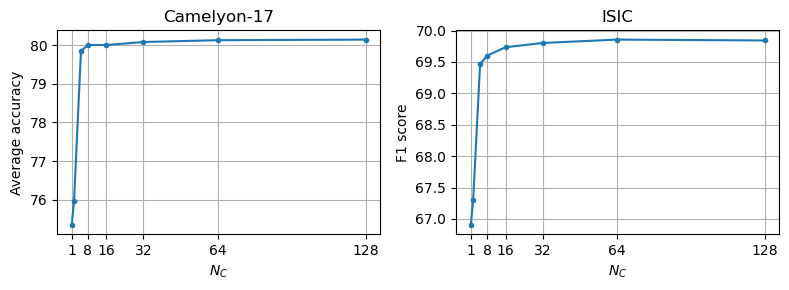}
\caption{
\nwbe{} performance vs $N_c$ for Camelyon-17 and ISIC datasets. Full mode. Performance is relativity insensitive to $N_c$ above $\sim 5$ examples per class.
}\label{fig:N_c}
\end{figure}

Table~\ref{tab:hyperparameters} shows hyperparameter settings for all datasets, where NW-specific hyperparameters are below the midline. 
For all models, we use pretrained ImageNet weights.
For $\lambda$, we perform a grid-search over the values \{0.01, 0.1, 1\}.
Fig.~\ref{fig:N_c} depicts \nwbe{} performance vs $N_c$ for Camelyon-17 and ISIC datasets. 
We find that performance is relativity insensitive to $N_c$ above $\sim 5$ examples per class.

In the original NW head paper~\cite{wang2023nwhead}, the authors experiment with a temperature (i.e. bandwidth) hyperparameter $\tau$. In this work, we set $\tau=1$ for all experiments.
The reason for this is that we optimize both the feature extractor and classifier end-to-end, and the kernel used in the classifier is dependent on the features that the feature extractor learns (unlike~\cite{kotelevskii2022nonparametric}, e.g.). 
Thus, we let the feature extractor to optimize the bandwidth on its own.
Note this same approach is taken in prior works~\cite{snell2017prototypical}.

\input{tables/table_hyperparameters}

\clearpage
\section{Table of Runtimes}
Table~\ref{tab:runtimes} shows approximate runtimes for various datasets during training and inference.
All experiments are performed on a GPU.
\input{tables/table_runtimes}

\section{ID vs. OOD Performance}
With most invariant learning methods, prior works observe a tradeoff between in-distribution and out-of-distribution generalization.
In Table~\ref{tab:id_vs_ood}, we show in-distribution (ID) and out-of-distribution (OOD) results for Camelyon-17, which provides an ID validation set.
We observe that there is a tradeoff between ID and OOD performance, similar to prior work.
\input{tables/table_id_vs_ood}

\clearpage
\section{Imbalanced ISIC Experiments}
\begin{figure}[h]
\centering
\includegraphics[width=\linewidth]{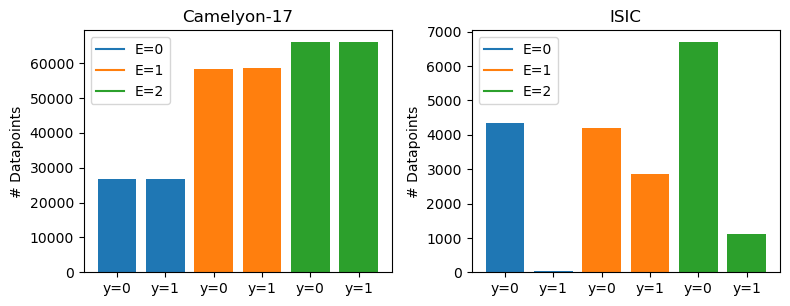}
\caption{
Number of datapoints separated by class for Camelyon-17 and ISIC datasets. 
There is significant label imbalance for the ISIC dataset.
}\label{fig:camelyon_isic_label_imbalance}
\end{figure}

As ISIC exhibits significant label imbalance where the positive class is much less represented than the negative class (see Fig.~\ref{fig:camelyon_isic_label_imbalance}), we experiment with an NW variant without support-set label balancing as an ablation.
For both variants, we set $N_c=8$.
To train the imbalanced variant, we sample a mini-batch support set by first sampling one image per class (to guarantee both classes are represented in the support at least once), and then sampling the rest of the images randomly from the dataset. 

To characterize the performance of both variants across class imbalances, we change the prevalance of $y=0$ in the test set by removing negative class images until the desired prevalence is achieved. 
Note the default prevalence is $\sim0.85$.
Then, we compute the accuracy over this manipulated test set (note, prior work has shown that F1-score is not a good metric for comparing classifiers with different label imbalances~\cite{brabec2020model}).

\begin{figure}[h]
\centering
\includegraphics[width=\linewidth]{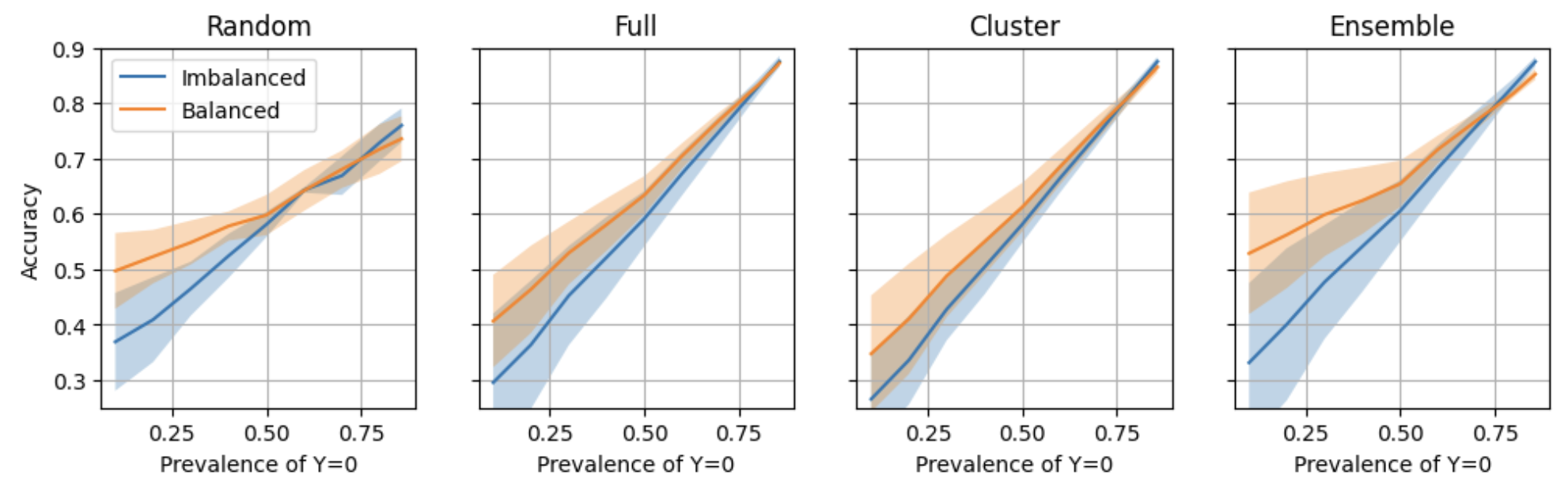}
\caption{
Accuracy of NW (imbalanced) and \nwb{} (balanced) models over varying prevalence of $y=0$ for ISIC dataset.
At low prevalence where the prevalence differs the most from training domains, we observe that model performance is higher for \nwb{}.
The default prevalence is 6705/7818 = 0.8576, which is the right-most value in the x-axis.
}\label{fig:isic_imbalance_experiments}
\end{figure}

Fig~\ref{fig:isic_imbalance_experiments} shows results.
We observe that at high prevalence where the proportion of negative classes matches in training and test, the imbalanced variant outperforms the balanced variant, whereas the opposite is true for low prevalence.
This makes sense because at high prevalence, the class imbalance is similar for the training and test domains; thus, a model which overpredicts the negative class is usually right.
On the other hand, this fails in test sets where the prevalence is flipped (i.e. low prevalence).
These results suggest that \nwb{} is a more robust classifier in the presence of label shift.

\clearpage
\section{Nearest-neighbor Inference Modes}
In addition to Random, Full, Cluster, and Ensemble inference modes, we additionally experiment with two inference modes based on nearest neighbors: k-NN and HNSW. 
HNSW (Hierarchical Navigable Small Worlds) is a fast approximate nearest neighbor algorithm~\cite{malkov2018efficient}. We choose $k=20$ based on prior work~\cite{kotelevskii2022nonparametric}. 
HNSW is about two times faster in total runtime on a GPU than full k-NN.%; we plan to add these findings to our computational results in Table 4.

Overall, we observe that k-NN and HNSW perform nearly identically for all the datasets and variants. 
Additionally, both modes perform better in terms of mean performance on Camelyon-17, and perform on par with the best-performing modes for ISIC (Cluster) and FMoW (Ensemble). 
However, they generally have higher variances across model runs. 
We suspect that less total samples used in the support (20 as compared to more than 1000 for Full) and the fact that not all classes are guaranteed to be represented in the support may lead to less stability across model runs.
\input{tables/table_results_knn_hnsw}

\section{Interpretability of NW Head}
\begin{figure}[h]
\centering
\includegraphics[width=\linewidth]{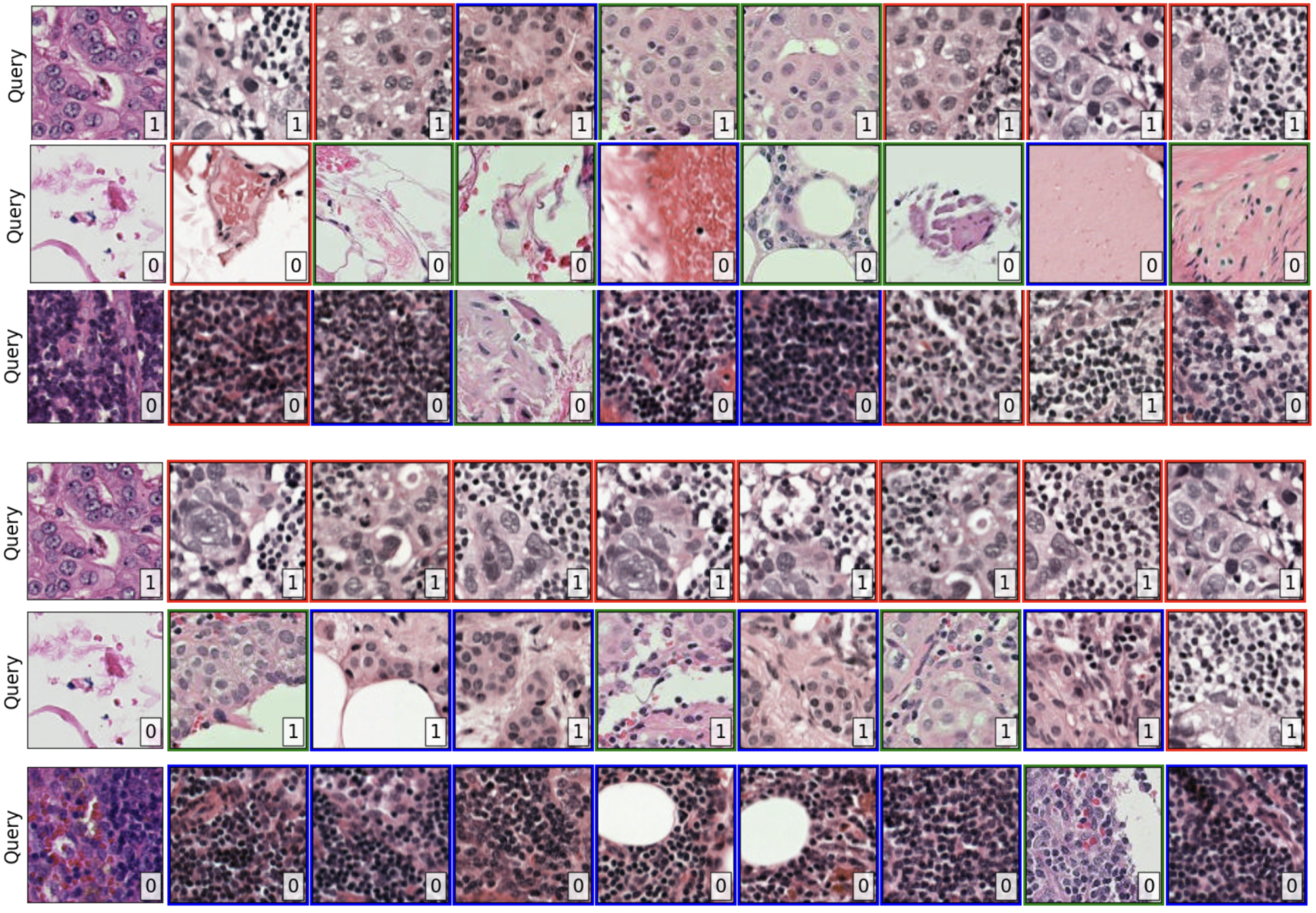}
\caption{
Visualization of 3 query images and their 8 nearest neighbors in the feature space, for both \nwbe{} (top) and \nwb{} (bottom). 
Labels are shown in the bottom right corner, and colored borders of neighbors indicate which of the 3 training environments the image comes from. For both variants, we observe visual similarity between the query images and the nearest neighbors. 
However, we observe the neighbors for \nwb{} tend to lack diversity in the environments from which they originate. 
In contrast, neighbors for \nwbe{} tend to come from a wider variety of environments, suggesting that \nwbe{} captures more invariant representations.
}\label{fig:nearest-neighbors}
\end{figure}

In this section, we provide both a visual and quantitative exploration of the interpretability capabilities of the NW head. 
In Fig.~\ref{fig:nearest-neighbors}, we show several query images and its 8 nearest neighbors in the feature space, for both \nwb{} and \nwbe{} variants. 
The colored border around each neighbor indicates which training environment the image comes from. 
Interestingly, we notice that the neighbors for \nwbe{} come from a variety of environments (note a variety of colored borders), while the neighbors for \nwb{} are less diverse. 

\begin{figure}[h]
\centering
\includegraphics[width=\linewidth]{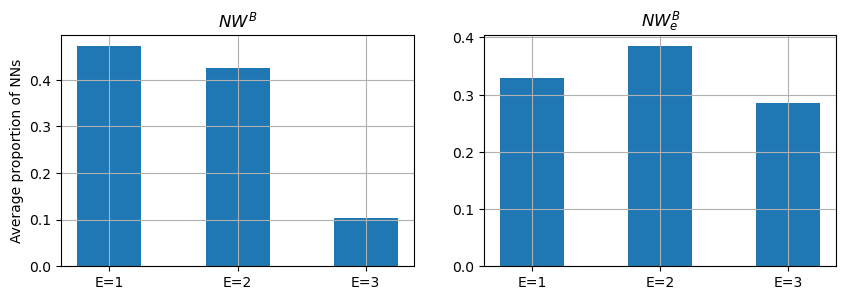}
\caption{
Normalized histogram of the environments from which the top 20 nearest neighbors originate in the training dataset for Camelyon-17, averaged over all queries in the test set. We observe a more balanced proportion for \nwbe{}, indicating that the model relies more evenly across all 3 environments to make its prediction, and further suggesting that representations are more invariant than \nwb{}. 
}\label{fig:average_proportion_nns}
\end{figure}

Fig.~\ref{fig:average_proportion_nns} quantifies this phenomenon, depicting a normalized histogram of the environments from which the top 20 nearest neighbors originate in the training dataset for Camelyon-17, averaged over all queries in the test set. 
From these results, it is clear that \nwbe{} leverages support images from a wider variety of environments to make its prediction, suggesting that it captures more invariant representations.
 
\label{sec:interpretability}

\end{document}

%% file: tables/table_datasets.tex
\begin{table*}[t]
    \centering
    \caption{Summary of Datasets.}
    \label{tab:datasets}
\begin{tabular*}{\textwidth}{@{\extracolsep{\fill}}llllllll@{}}
\toprule
             \textit{Dataset}  & \textit{\# Classes} & \textit{Env} & \textit{\# Envs}           & \textit{Architecture}                  & \textit{Metric}      \\ \midrule
Camelyon-17   &  2     & Hospital & 3      & DenseNet-121             & 
 Average acc.  \\
ISIC   &  2 &  Hospital & 3   & ResNet-50            & F1-score   \\
FMoW     & 62   & Region & 5    & DenseNet-121             & Worst-region acc.   \\\bottomrule
\end{tabular*}
\end{table*}

%% file: tables/table_results.tex
\begin{table*}[t]

    \centering
    \caption{
    % Average accuracy for Camelyon-17, worst-group accuracy for FMoW, and F1-score for ISIC (\%). 
    Metric average $\pm$ standard deviation for all datasets (\%).
    Higher is better. 
    \textbf{Bold} is best and \underline{underline} is second-best. Implicit / Explicit.}
    \label{tab:results}
\begin{tabular*}{\textwidth}{@{\extracolsep{\fill}}lllll@{}}
\toprule
\textit{Algorithm}                            & \textit{Camelyon-17}                                  & \textit{ISIC}                                                & \textit{FMoW}      \\ \midrule
ERM~\cite{vapnik1999erm}                      & 70.3{\tiny$\pm$6.4}                                   & 58.2{\tiny$\pm$2.9}                                          & 32.6{\tiny$\pm$1.6} \\
IRM~\cite{arjovsky2019irm}                    & 70.9{\tiny$\pm$6.8}                                   & 57.9{\tiny$\pm$1.0}                                          & 31.3{\tiny$\pm$1.2}  \\
CORAL~\cite{sun2016coral}                     & 72.4{\tiny$\pm$4.4}                                   & 59.1{\tiny$\pm$2.2}                                          & 31.7{\tiny$\pm$1.0} \\
Fish~\cite{shi2021fish}                       & 74.7{\tiny$\pm$7e-2}                                  & 64.4{\tiny$\pm$1.7}                                          & 34.6{\tiny$\pm$0.0}  \\
LISA~\cite{yao2022lisa}                       & 77.1{\tiny$\pm$6.5}                                   & 64.8{\tiny$\pm$2.3}                                          & 35.5{\tiny$\pm$1.8} \\
CLOvE~\cite{wald2021oodcalibration}           & \underline{79.9}{\tiny$\pm$3.9}                       & 66.2{\tiny$\pm$2.2}                                          & \textbf{40.1}{\tiny$\pm$0.6}  \\\midrule
\nwb, Random                                  & 71.7{\tiny$\pm$5.3}                                   & 56.7{\tiny$\pm$1.4}                                          & 31.1{\tiny$\pm$0.8}  \\
\nwb, Full                                    & 72.0{\tiny$\pm$6.7}                                   & 61.9{\tiny$\pm$3.5}                                          & 31.6{\tiny$\pm$0.9} \\ 
\nwb, Cluster                                 & 70.6{\tiny$\pm$6.9}                                   & 61.4{\tiny$\pm$2.3}                                          & 31.3{\tiny$\pm$0.9}      \\
\nwb, Ensemble                                & 71.9{\tiny$\pm$6.0}                                   & 63.9{\tiny$\pm$3.8}                                          & 32.2{\tiny$\pm$1.0}     \\
\nwb, Probe                                   & 69.2{\tiny$\pm$7.4}                                   & 59.7{\tiny$\pm$2.5}                                          & 29.9{\tiny$\pm$1.5}      \\\midrule
\nwbe, Random                                 & 74.8{\tiny$\pm$8.4} / 75.3{\tiny$\pm$3.2}             & 57.5{\tiny$\pm$1.9} / 55.0{\tiny$\pm$0.9}                    & 31.2{\tiny$\pm$0.7} / 30.9{\tiny$\pm$0.5}\\
\nwbe, Full                                   & \textbf{80.0}{\tiny$\pm$2.7} / 79.7{\tiny$\pm$1.9}    & 69.6{\tiny$\pm$2.3} / 70.0{\tiny$\pm$1.0}        & 35.0{\tiny$\pm$0.7} / 34.6{\tiny$\pm$0.4}\\ 
\nwbe, Cluster                                & 78.6{\tiny$\pm$2.5} / 79.0{\tiny$\pm$1.4}             & \textbf{71.1}{\tiny$\pm$1.7} /  \underline{71.0}{\tiny$\pm$1.0}          & 33.9{\tiny$\pm$0.6} / 34.0{\tiny$\pm$0.3} \\
\nwbe, Ensemble                               & 79.5{\tiny$\pm$2.6} / 79.6{\tiny$\pm$1.9}             & 69.5{\tiny$\pm$2.2} / 69.8{\tiny$\pm$0.8}                    & 37.8{\tiny$\pm$0.9} / \underline{38.2}{\tiny$\pm$0.4} \\
\nwbe, Probe                                  & 75.3{\tiny$\pm$7.3} / 75.8{\tiny$\pm$8.3}             & 61.4{\tiny$\pm$3.1} / 63.4{\tiny$\pm$2.8}                    & 33.9{\tiny$\pm$1.5} / 32.7{\tiny$\pm$1.4}  \\\bottomrule
\end{tabular*}
\end{table*}

%% file: tables/table_hyperparameters.tex
\begin{table*}[h]

    \centering
    \caption{
    % Average accuracy for Camelyon-17, worst-group accuracy for FMoW, and F1-score for ISIC (\%). 
    Hyperparameter settings for various datasets.}
    \label{tab:hyperparameters}
\begin{tabular*}{\textwidth}{@{\extracolsep{\fill}}lllll@{}}
\toprule
\textit{Hyperparameter}                            & \textit{Camelyon-17}                                  & \textit{ISIC}                                                & \textit{FMoW}      \\ \midrule
Learning rate                      & 1e-4  & 5e-5  & 1e-4 \\
Weight decay                    & 1e-4      & 0 & 1e-2  \\
Scheduler                     & None & None & StepLR \\
Batch size                       & 32 & 8 & 8  \\
Architecture                       & DenseNet-121  & ResNet-50 & DenseNet-121 \\
Optimizer           & SGD & Adam & Adam  \\
Maximum Epoch       & 10    & 5 & 60  \\\midrule
$N_q$  & 8 & 8 & 8 \\
$N_c$  & 8 & 8 & 1 \\
$N_s$ & $N_c\times2 = 16$ & $N_c\times2 = 16$ & $N_c\times 62 = 62$ \\
$\lambda$ &0.01 &0.01& 0.1 \\
$k$ &3 &3& 3 \\
\bottomrule
\end{tabular*}
\end{table*}

%% file: tables/table_runtimes.tex
\begin{table*}[h]

    \centering
    \caption{
    % Average accuracy for Camelyon-17, worst-group accuracy for FMoW, and F1-score for ISIC (\%). 
    Approximate runtimes for various algorithms. Training time is time to complete maximum epochs as specified in Table~\ref{tab:hyperparameters}, and does not include validation. 
    Inference time is time to evaluate the entire test set.
    Averaged over all training runs.}
    \label{tab:runtimes}
\begin{tabular*}{\textwidth}{@{\extracolsep{\fill}}lllll@{}}
\toprule
  & \textit{Algorithm} & \textit{Camelyon-17} & \textit{ISIC} & \textit{FMoW} \\ \midrule
Training & ERM         & 7 hr & 1 hr & 22 hr  \\
&NW                    & 14 hr & 2 hr & 40 hr  \\\midrule
Inference & ERM        & 10 min & 2 min & 10 min  \\
&NW, Random            & 15 min & 3 min & 20 min  \\
&NW, Full              & 2 hr & 15 min & 1 hr  \\
&NW, Ensemble          & 2 hr & 15 min & 1 hr  \\
&NW, Cluster           & 2.2 hr & 17 min & 1.1 hr \\
&NW, Probe             & 10 min & 2 min & 10 min \\
\bottomrule
\end{tabular*}
\end{table*}

%% file: tables/table_id_vs_ood.tex
\begin{table*}[h]
    \centering
    \caption{
    In-distribution vs. out-of-distribution generalization performance on Camelyon-17.}
    \label{tab:id_vs_ood}
\begin{tabular*}{\textwidth}{@{\extracolsep{\fill}}lll@{}}
\toprule
                            & \textit{In-distribution}                                  & \textit{Out-of-distribution}  \\ \midrule
ERM                      & 93.2{\tiny$\pm$5.2}   & 70.3{\tiny$\pm$6.4}   \\
\nwb{}, Full                    & 96.1{\tiny$\pm$1.0}  & 72.0{\tiny$\pm$6.7}   \\
\nwbe{}, Full                     & 92.8{\tiny$\pm$2.0} & 80.0{\tiny$\pm$2.7} \\
\bottomrule
\end{tabular*}
\end{table*}

%% file: tables/table_results_knn_hnsw.tex
\begin{table*}[h]

    \centering
    \caption{
    % Average accuracy for Camelyon-17, worst-group accuracy for FMoW, and F1-score for ISIC (\%). 
    Metric average $\pm$ standard deviation for all datasets (\%).
    Higher is better. Implicit / Explicit.}
    \label{tab:nn}
\begin{tabular*}{\textwidth}{@{\extracolsep{\fill}}lllll@{}}
\toprule
\textit{Algorithm}                            & \textit{Camelyon-17}                                  & \textit{ISIC}                                                & \textit{FMoW}      \\ \midrule
\nwb, Full                                    & 72.0{\tiny$\pm$6.7}                                   & 61.9{\tiny$\pm$3.5}                                          & 31.6{\tiny$\pm$0.9} \\ 
\nwb, Cluster                                 & 70.6{\tiny$\pm$6.9}                                   & 61.4{\tiny$\pm$2.3}                                          & 31.3{\tiny$\pm$0.9}      \\
\nwb, Ensemble                                & 71.9{\tiny$\pm$6.0}                                   & 63.9{\tiny$\pm$3.8}                                          & 32.2{\tiny$\pm$1.0}     \\
\nwb, k-NN                                & 72.5{\tiny$\pm$3.2}                                   & 64.2{\tiny$\pm$2.6}                                          & 32.5{\tiny$\pm$1.4}     \\
\nwb, HNSW                                & 72.5{\tiny$\pm$3.2}                                   & 64.2{\tiny$\pm$2.6}                                          & 32.5{\tiny$\pm$1.4}     \\\midrule
%%%%%%%%%
\nwbe, Full                                   & 80.0{\tiny$\pm$2.7} / 79.7{\tiny$\pm$1.9}    & 69.6{\tiny$\pm$2.3} / 70.0{\tiny$\pm$1.0}        & 35.0{\tiny$\pm$0.7} / 34.6{\tiny$\pm$0.4}\\ 
\nwbe, Cluster                                & 78.6{\tiny$\pm$2.5} / 79.0{\tiny$\pm$1.4}             & 71.1{\tiny$\pm$1.7} /  71.0{\tiny$\pm$1.0}          & 33.9{\tiny$\pm$0.6} / 34.0{\tiny$\pm$0.3} \\
\nwbe, Ensemble                               & 79.5{\tiny$\pm$2.6} / 79.6{\tiny$\pm$1.9}             & 69.5{\tiny$\pm$2.2} / 69.8{\tiny$\pm$0.8}                    & 37.8{\tiny$\pm$0.9} / 38.2{\tiny$\pm$0.4} \\
\nwbe, k-NN                               & 80.9{\tiny$\pm$10.3} / 80.2{\tiny$\pm$4.9}             & 70.7{\tiny$\pm$7.4} / 71.0{\tiny$\pm$7.2}                    & 37.9{\tiny$\pm$2.3} / 38.3{\tiny$\pm$1.1} \\
\nwbe, HNSW                               & 80.9{\tiny$\pm$10.3} / 80.2{\tiny$\pm$5.0}             & 70.7{\tiny$\pm$7.4} / 71.0{\tiny$\pm$7.2}                    & 37.9{\tiny$\pm$2.2} / 38.2{\tiny$\pm$1.1} \\\bottomrule
\end{tabular*}
\end{table*}

%% file: main.bbl
\begin{thebibliography}{71}
\providecommand{\natexlab}[1]{#1}
\providecommand{\url}[1]{\texttt{#1}}
\expandafter\ifx\csname urlstyle\endcsname\relax
  \providecommand{\doi}[1]{doi: #1}\else
  \providecommand{\doi}{doi: \begingroup \urlstyle{rm}\Url}\fi

\bibitem[Arjovsky et~al.(2019)Arjovsky, Bottou, Gulrajani, and Lopez-Paz]{arjovsky2019irm}
Martin Arjovsky, Léon Bottou, Ishaan Gulrajani, and David Lopez-Paz.
\newblock Invariant risk minimization, 2019.
\newblock URL \url{https://arxiv.org/abs/1907.02893}.

\bibitem[Bishop(2006)]{bishop2006mlbook}
Christopher~M. Bishop.
\newblock \emph{Pattern Recognition and Machine Learning (Information Science and Statistics)}.
\newblock Springer-Verlag, Berlin, Heidelberg, 2006.
\newblock ISBN 0387310738.

\bibitem[Brabec et~al.(2020)Brabec, Komárek, Franc, and Machlica]{brabec2020model}
Jan Brabec, Tomáš Komárek, Vojtěch Franc, and Lukáš Machlica.
\newblock On model evaluation under non-constant class imbalance, 2020.

\bibitem[Bándi et~al.(2019)Bándi, Geessink, Manson, Van~Dijk, Balkenhol, Hermsen, Ehteshami~Bejnordi, Lee, Paeng, Zhong, Li, Zanjani, Zinger, Fukuta, Komura, Ovtcharov, Cheng, Zeng, Thagaard, Dahl, Lin, Chen, Jacobsson, Hedlund, Çetin, Halıcı, Jackson, Chen, Both, Franke, Küsters-Vandevelde, Vreuls, Bult, van Ginneken, van~der Laak, and Litjens]{bandi2019camelyon}
Péter Bándi, Oscar Geessink, Quirine Manson, Marcory Van~Dijk, Maschenka Balkenhol, Meyke Hermsen, Babak Ehteshami~Bejnordi, Byungjae Lee, Kyunghyun Paeng, Aoxiao Zhong, Quanzheng Li, Farhad~Ghazvinian Zanjani, Svitlana Zinger, Keisuke Fukuta, Daisuke Komura, Vlado Ovtcharov, Shenghua Cheng, Shaoqun Zeng, Jeppe Thagaard, Anders~B. Dahl, Huangjing Lin, Hao Chen, Ludwig Jacobsson, Martin Hedlund, Melih Çetin, Eren Halıcı, Hunter Jackson, Richard Chen, Fabian Both, Jörg Franke, Heidi Küsters-Vandevelde, Willem Vreuls, Peter Bult, Bram van Ginneken, Jeroen van~der Laak, and Geert Litjens.
\newblock From detection of individual metastases to classification of lymph node status at the patient level: The camelyon17 challenge.
\newblock \emph{IEEE Transactions on Medical Imaging}, 38\penalty0 (2):\penalty0 550--560, 2019.
\newblock \doi{10.1109/TMI.2018.2867350}.

\bibitem[Chevalley et~al.(2022)Chevalley, Bunne, Krause, and Bauer]{chevalley2022invariant}
Mathieu Chevalley, Charlotte Bunne, Andreas Krause, and Stefan Bauer.
\newblock Invariant causal mechanisms through distribution matching, 2022.

\bibitem[Christie et~al.(2018)Christie, Fendley, Wilson, and Mukherjee]{christie2018functional}
Gordon Christie, Neil Fendley, James Wilson, and Ryan Mukherjee.
\newblock Functional map of the world, 2018.

\bibitem[Codella et~al.(2019)Codella, Rotemberg, Tschandl, Celebi, Dusza, Gutman, Helba, Kalloo, Liopyris, Marchetti, et~al.]{codella2019skin}
Noel Codella, Veronica Rotemberg, Philipp Tschandl, M~Emre Celebi, Stephen Dusza, David Gutman, Brian Helba, Aadi Kalloo, Konstantinos Liopyris, Michael Marchetti, et~al.
\newblock Skin lesion analysis toward melanoma detection 2018: A challenge hosted by the international skin imaging collaboration (isic).
\newblock Eprint \href{http://arxiv.org/abs/1902.03368}{arXiv:1902.03368}, 2019.

\bibitem[Codella et~al.(2018)Codella, Gutman, Celebi, Helba, Marchetti, Dusza, Kalloo, Liopyris, Mishra, Kittler, et~al.]{codella2018skin}
Noel~CF Codella, David Gutman, M~Emre Celebi, Brian Helba, Michael~A Marchetti, Stephen~W Dusza, Aadi Kalloo, Konstantinos Liopyris, Nabin Mishra, Harald Kittler, et~al.
\newblock Skin lesion analysis toward melanoma detection: A challenge at the 2017 international symposium on biomedical imaging (isbi), hosted by the international skin imaging collaboration (isic).
\newblock In \emph{Proceedings of ISBI}, pages 168--172. IEEE, 2018.

\bibitem[Combalia et~al.(2019)Combalia, Codella, Rotemberg, Helba, Vilaplana, Reiter, Carrera, Barreiro, Halpern, Puig, et~al.]{combalia2019bcn20000}
Marc Combalia, Noel~CF Codella, Veronica Rotemberg, Brian Helba, Veronica Vilaplana, Ofer Reiter, Cristina Carrera, Alicia Barreiro, Allan~C Halpern, Susana Puig, et~al.
\newblock Bcn20000: Dermoscopic lesions in the wild.
\newblock Eprint \href{http://arxiv.org/abs/1908.02288}{arXiv:1908.02288}, 2019.

\bibitem[Damianou and Lawrence(2013)]{damianou2013gps}
Andreas Damianou and Neil~D. Lawrence.
\newblock Deep gaussian processes.
\newblock In Carlos~M. Carvalho and Pradeep Ravikumar, editors, \emph{Proceedings of the Sixteenth International Conference on Artificial Intelligence and Statistics}, volume~31 of \emph{Proceedings of Machine Learning Research}, pages 207--215, Scottsdale, Arizona, USA, 29 Apr--01 May 2013. PMLR.
\newblock URL \url{https://proceedings.mlr.press/v31/damianou13a.html}.

\bibitem[Dosovitskiy et~al.(2020)Dosovitskiy, Beyer, Kolesnikov, Weissenborn, Zhai, Unterthiner, Dehghani, Minderer, Heigold, Gelly, Uszkoreit, and Houlsby]{dosovitskiy2020vit}
Alexey Dosovitskiy, Lucas Beyer, Alexander Kolesnikov, Dirk Weissenborn, Xiaohua Zhai, Thomas Unterthiner, Mostafa Dehghani, Matthias Minderer, Georg Heigold, Sylvain Gelly, Jakob Uszkoreit, and Neil Houlsby.
\newblock An image is worth 16x16 words: Transformers for image recognition at scale, 2020.
\newblock URL \url{https://arxiv.org/abs/2010.11929}.

\bibitem[Dou et~al.(2019)Dou, Coelho~de Castro, Kamnitsas, and Glocker]{dou2019semantic}
Qi~Dou, Daniel Coelho~de Castro, Konstantinos Kamnitsas, and Ben Glocker.
\newblock Domain generalization via model-agnostic learning of semantic features.
\newblock In H.~Wallach, H.~Larochelle, A.~Beygelzimer, F.~d\textquotesingle Alch\'{e}-Buc, E.~Fox, and R.~Garnett, editors, \emph{Advances in Neural Information Processing Systems}, volume~32. Curran Associates, Inc., 2019.
\newblock URL \url{https://proceedings.neurips.cc/paper_files/paper/2019/file/2974788b53f73e7950e8aa49f3a306db-Paper.pdf}.

\bibitem[Fakoor et~al.(2020)Fakoor, Chaudhari, Mueller, and Smola]{fakoor2020trade}
Rasool Fakoor, Pratik Chaudhari, Jonas Mueller, and Alexander~J. Smola.
\newblock Trade: Transformers for density estimation, 2020.
\newblock URL \url{https://arxiv.org/abs/2004.02441}.

\bibitem[Ganin et~al.(2016)Ganin, Ustinova, Ajakan, Germain, Larochelle, Laviolette, Marchand, and Lempitsky]{ganin2016domainadversarial}
Yaroslav Ganin, Evgeniya Ustinova, Hana Ajakan, Pascal Germain, Hugo Larochelle, François Laviolette, Mario Marchand, and Victor Lempitsky.
\newblock Domain-adversarial training of neural networks, 2016.

\bibitem[Ghosh and Nag(2001)]{ghosh2001overviewrbfs}
J.~Ghosh and A.~Nag.
\newblock \emph{An Overview of Radial Basis Function Networks}, pages 1--36.
\newblock Physica-Verlag HD, Heidelberg, 2001.
\newblock ISBN 978-3-7908-1826-0.
\newblock \doi{10.1007/978-3-7908-1826-0_1}.
\newblock URL \url{https://doi.org/10.1007/978-3-7908-1826-0_1}.

\bibitem[Gulrajani and Lopez-Paz(2020)]{gulrajani2020search}
Ishaan Gulrajani and David Lopez-Paz.
\newblock In search of lost domain generalization, 2020.

\bibitem[Gutman et~al.(2016)Gutman, Codella, Celebi, Helba, Marchetti, Mishra, and Halpern]{gutman2016skin}
David Gutman, Noel~CF Codella, Emre Celebi, Brian Helba, Michael Marchetti, Nabin Mishra, and Allan Halpern.
\newblock Skin lesion analysis toward melanoma detection: A challenge at the international symposium on biomedical imaging (isbi) 2016, hosted by the international skin imaging collaboration (isic).
\newblock Eprint \href{http://arxiv.org/abs/1605.01397}{arXiv:1605.01397}, 2016.

\bibitem[Heinze-Deml and Meinshausen(2019)]{heinzedeml2019conditional}
Christina Heinze-Deml and Nicolai Meinshausen.
\newblock Conditional variance penalties and domain shift robustness, 2019.

\bibitem[Jaegle et~al.(2021)Jaegle, Gimeno, Brock, Zisserman, Vinyals, and Carreira]{jaegle2021perceiver}
Andrew Jaegle, Felix Gimeno, Andrew Brock, Andrew Zisserman, Oriol Vinyals, and Joao Carreira.
\newblock Perceiver: General perception with iterative attention, 2021.
\newblock URL \url{https://arxiv.org/abs/2103.03206}.

\bibitem[Jiang and Veitch(2022)]{jiang2022transportable}
Yibo Jiang and Victor Veitch.
\newblock Invariant and transportable representations for anti-causal domain shifts, 2022.
\newblock URL \url{https://arxiv.org/abs/2207.01603}.

\bibitem[Kaddour et~al.(2022)Kaddour, Lynch, Liu, Kusner, and Silva]{kaddour2022causalsurvey}
Jean Kaddour, Aengus Lynch, Qi~Liu, Matt~J. Kusner, and Ricardo Silva.
\newblock Causal machine learning: A survey and open problems, 2022.
\newblock URL \url{https://arxiv.org/abs/2206.15475}.

\bibitem[Kamath et~al.(2021)Kamath, Tangella, Sutherland, and Srebro]{kamath2021does}
Pritish Kamath, Akilesh Tangella, Danica~J. Sutherland, and Nathan Srebro.
\newblock Does invariant risk minimization capture invariance?, 2021.

\bibitem[Koh et~al.(2021)Koh, Sagawa, Marklund, Xie, Zhang, Balsubramani, Hu, Yasunaga, Phillips, Gao, Lee, David, Stavness, Guo, Earnshaw, Haque, Beery, Leskovec, Kundaje, Pierson, Levine, Finn, and Liang]{koh2021wilds}
Pang~Wei Koh, Shiori Sagawa, Henrik Marklund, Sang~Michael Xie, Marvin Zhang, Akshay Balsubramani, Weihua Hu, Michihiro Yasunaga, Richard~Lanas Phillips, Irena Gao, Tony Lee, Etienne David, Ian Stavness, Wei Guo, Berton~A. Earnshaw, Imran~S. Haque, Sara Beery, Jure Leskovec, Anshul Kundaje, Emma Pierson, Sergey Levine, Chelsea Finn, and Percy Liang.
\newblock Wilds: A benchmark of in-the-wild distribution shifts, 2021.

\bibitem[Kossen et~al.(2021)Kossen, Band, Lyle, Gomez, Rainforth, and Gal]{kossen2021npt}
Jannik Kossen, Neil Band, Clare Lyle, Aidan~N. Gomez, Tom Rainforth, and Yarin Gal.
\newblock Self-attention between datapoints: Going beyond individual input-output pairs in deep learning, 2021.
\newblock URL \url{https://arxiv.org/abs/2106.02584}.

\bibitem[Kotelevskii et~al.(2022)Kotelevskii, Artemenkov, Fedyanin, Noskov, Fishkov, Shelmanov, Vazhentsev, Petiushko, and Panov]{kotelevskii2022nonparametric}
Nikita Kotelevskii, Aleksandr Artemenkov, Kirill Fedyanin, Fedor Noskov, Alexander Fishkov, Artem Shelmanov, Artem Vazhentsev, Aleksandr Petiushko, and Maxim Panov.
\newblock Nonparametric uncertainty quantification for single deterministic neural network, 2022.

\bibitem[Koyama and Yamaguchi(2021)]{koyama2021invariance}
Masanori Koyama and Shoichiro Yamaguchi.
\newblock When is invariance useful in an out-of-distribution generalization problem ?, 2021.

\bibitem[Krueger et~al.(2021)Krueger, Caballero, Jacobsen, Zhang, Binas, Zhang, Priol, and Courville]{krueger2021outofdistribution}
David Krueger, Ethan Caballero, Joern-Henrik Jacobsen, Amy Zhang, Jonathan Binas, Dinghuai Zhang, Remi~Le Priol, and Aaron Courville.
\newblock Out-of-distribution generalization via risk extrapolation (rex), 2021.

\bibitem[Kumar et~al.(2018)Kumar, Sarawagi, and Jain]{kumar2018mmce}
Aviral Kumar, Sunita Sarawagi, and Ujjwal Jain.
\newblock Trainable calibration measures for neural networks from kernel mean embeddings.
\newblock In Jennifer Dy and Andreas Krause, editors, \emph{Proceedings of the 35th International Conference on Machine Learning}, volume~80 of \emph{Proceedings of Machine Learning Research}, pages 2805--2814. PMLR, 10--15 Jul 2018.
\newblock URL \url{https://proceedings.mlr.press/v80/kumar18a.html}.

\bibitem[Li et~al.(2017)Li, Yang, Song, and Hospedales]{li2017learning}
Da~Li, Yongxin Yang, Yi-Zhe Song, and Timothy~M. Hospedales.
\newblock Learning to generalize: Meta-learning for domain generalization, 2017.

\bibitem[Li et~al.(2018)Li, Pan, Wang, and Kot]{li2018adversarial}
Haoliang Li, Sinno~Jialin Pan, Shiqi Wang, and Alex~C. Kot.
\newblock Domain generalization with adversarial feature learning.
\newblock In \emph{2018 IEEE/CVF Conference on Computer Vision and Pattern Recognition}, pages 5400--5409, 2018.
\newblock \doi{10.1109/CVPR.2018.00566}.

\bibitem[Lipton et~al.(2018)Lipton, Wang, and Smola]{lipton2018detecting}
Zachary~C. Lipton, Yu-Xiang Wang, and Alex Smola.
\newblock Detecting and correcting for label shift with black box predictors, 2018.

\bibitem[Long et~al.(2015)Long, Cao, Wang, and Jordan]{long2015learning}
Mingsheng Long, Yue Cao, Jianmin Wang, and Michael~I. Jordan.
\newblock Learning transferable features with deep adaptation networks, 2015.

\bibitem[Malkov and Yashunin(2018)]{malkov2018efficient}
Yu.~A. Malkov and D.~A. Yashunin.
\newblock Efficient and robust approximate nearest neighbor search using hierarchical navigable small world graphs, 2018.

\bibitem[Mansilla et~al.(2021)Mansilla, Echeveste, Milone, and Ferrante]{mansilla2021domain}
Lucas Mansilla, Rodrigo Echeveste, Diego~H. Milone, and Enzo Ferrante.
\newblock Domain generalization via gradient surgery, 2021.

\bibitem[Nadaraya(1964)]{nadaraya1964}
E.~A. Nadaraya.
\newblock On estimating regression.
\newblock \emph{Theory of Probability \& Its Applications}, 9\penalty0 (1):\penalty0 141--142, 1964.
\newblock \doi{10.1137/1109020}.
\newblock URL \url{https://doi.org/10.1137/1109020}.

\bibitem[Nguyen et~al.(2021)Nguyen, Chen, and Lee]{nguyen2021krr}
Timothy Nguyen, Zhourong Chen, and Jaehoon Lee.
\newblock Dataset meta-learning from kernel ridge-regression, 2021.

\bibitem[Papernot and McDaniel(2018)]{papernot2018deepknn}
Nicolas Papernot and Patrick McDaniel.
\newblock Deep k-nearest neighbors: Towards confident, interpretable and robust deep learning, 2018.
\newblock URL \url{https://arxiv.org/abs/1803.04765}.

\bibitem[Parmar et~al.(2018)Parmar, Vaswani, Uszkoreit, Kaiser, Shazeer, Ku, and Tran]{parmar2018imagetransformer}
Niki Parmar, Ashish Vaswani, Jakob Uszkoreit, Łukasz Kaiser, Noam Shazeer, Alexander Ku, and Dustin Tran.
\newblock Image transformer, 2018.
\newblock URL \url{https://arxiv.org/abs/1802.05751}.

\bibitem[Pearl(2009)]{pearl2009causality}
Judea Pearl.
\newblock \emph{Causality}.
\newblock Cambridge University Press, 2 edition, 2009.
\newblock \doi{10.1017/CBO9780511803161}.

\bibitem[Peters et~al.(2015)Peters, Bühlmann, and Meinshausen]{peters2015causal}
Jonas Peters, Peter Bühlmann, and Nicolai Meinshausen.
\newblock Causal inference using invariant prediction: identification and confidence intervals, 2015.

\bibitem[Rojas-Carulla et~al.(2018)Rojas-Carulla, Schölkopf, Turner, and Peters]{rojascarulla2018invariant}
Mateo Rojas-Carulla, Bernhard Schölkopf, Richard Turner, and Jonas Peters.
\newblock Invariant models for causal transfer learning, 2018.

\bibitem[Rosenfeld et~al.(2021)Rosenfeld, Ravikumar, and Risteski]{rosenfeld2021risksofirm}
Elan Rosenfeld, Pradeep Ravikumar, and Andrej Risteski.
\newblock The risks of invariant risk minimization, 2021.

\bibitem[Rotemberg et~al.(2021)Rotemberg, Kurtansky, Betz-Stablein, Caffery, Chousakos, Codella, Combalia, Dusza, Guitera, Gutman, et~al.]{rotemberg2021patient}
Veronica Rotemberg, Nicholas Kurtansky, Brigid Betz-Stablein, Liam Caffery, Emmanouil Chousakos, Noel Codella, Marc Combalia, Stephen Dusza, Pascale Guitera, David Gutman, et~al.
\newblock A patient-centric dataset of images and metadata for identifying melanomas using clinical context.
\newblock \emph{Scientific data}, 8\penalty0 (1):\penalty0 1--8, 2021.

\bibitem[Schoelkopf et~al.(2012)Schoelkopf, Janzing, Peters, Sgouritsa, Zhang, and Mooij]{schoelkopf2012causal}
Bernhard Schoelkopf, Dominik Janzing, Jonas Peters, Eleni Sgouritsa, Kun Zhang, and Joris Mooij.
\newblock On causal and anticausal learning, 2012.

\bibitem[Scope et~al.(2009)Scope, Marghoob, Chen, Lieb, Weinstock, Halpern, and Group]{scope2009dermoscopic}
A~Scope, AA~Marghoob, CS~Chen, JA~Lieb, MA~Weinstock, AC~Halpern, and SONIC~Study Group.
\newblock Dermoscopic patterns and subclinical melanocytic nests in normal-appearing skin.
\newblock \emph{British Journal of Dermatology}, 160\penalty0 (6):\penalty0 1318--1321, 2009.

\bibitem[Shi et~al.(2021)Shi, Seely, Torr, Siddharth, Hannun, Usunier, and Synnaeve]{shi2021fish}
Yuge Shi, Jeffrey Seely, Philip H.~S. Torr, N.~Siddharth, Awni Hannun, Nicolas Usunier, and Gabriel Synnaeve.
\newblock Gradient matching for domain generalization, 2021.

\bibitem[Snell et~al.(2017)Snell, Swersky, and Zemel]{snell2017prototypical}
Jake Snell, Kevin Swersky, and Richard~S. Zemel.
\newblock Prototypical networks for few-shot learning, 2017.
\newblock URL \url{https://arxiv.org/abs/1703.05175}.

\bibitem[Sun and Saenko(2016)]{sun2016coral}
Baochen Sun and Kate Saenko.
\newblock Deep coral: Correlation alignment for deep domain adaptation, 2016.

\bibitem[Taesiri et~al.(2022)Taesiri, Nguyen, and Nguyen]{taesiri2022visual}
Mohammad~Reza Taesiri, Giang Nguyen, and Anh Nguyen.
\newblock Visual correspondence-based explanations improve {AI} robustness and human-{AI} team accuracy.
\newblock In Alice~H. Oh, Alekh Agarwal, Danielle Belgrave, and Kyunghyun Cho, editors, \emph{Advances in Neural Information Processing Systems}, 2022.
\newblock URL \url{https://openreview.net/forum?id=UavQ9HYye6n}.

\bibitem[Tschandl et~al.(2018)Tschandl, Rosendahl, and Kittler]{tschandl2018ham10000}
Philipp Tschandl, Cliff Rosendahl, and Harald Kittler.
\newblock The ham10000 dataset, a large collection of multi-source dermatoscopic images of common pigmented skin lesions.
\newblock \emph{Scientific data}, 5\penalty0 (1):\penalty0 1--9, 2018.

\bibitem[Tzeng et~al.(2014)Tzeng, Hoffman, Zhang, Saenko, and Darrell]{tzeng2014deep}
Eric Tzeng, Judy Hoffman, Ning Zhang, Kate Saenko, and Trevor Darrell.
\newblock Deep domain confusion: Maximizing for domain invariance, 2014.

\bibitem[Vapnik(1999)]{vapnik1999erm}
V.N. Vapnik.
\newblock An overview of statistical learning theory.
\newblock \emph{IEEE Transactions on Neural Networks}, 10\penalty0 (5):\penalty0 988--999, 1999.
\newblock \doi{10.1109/72.788640}.

\bibitem[Vaswani et~al.(2017)Vaswani, Shazeer, Parmar, Uszkoreit, Jones, Gomez, Kaiser, and Polosukhin]{vaswani2017attention}
Ashish Vaswani, Noam Shazeer, Niki Parmar, Jakob Uszkoreit, Llion Jones, Aidan~N. Gomez, Lukasz Kaiser, and Illia Polosukhin.
\newblock Attention is all you need, 2017.
\newblock URL \url{https://arxiv.org/abs/1706.03762}.

\bibitem[Wald et~al.(2021)Wald, Feder, Greenfeld, and Shalit]{wald2021oodcalibration}
Yoav Wald, Amir Feder, Daniel Greenfeld, and Uri Shalit.
\newblock On calibration and out-of-domain generalization, 2021.
\newblock URL \url{https://arxiv.org/abs/2102.10395}.

\bibitem[Wang and Sabuncu(2023)]{wang2023nwhead}
Alan~Q. Wang and Mert~R. Sabuncu.
\newblock A flexible nadaraya-watson head can offer explainable and calibrated classification.
\newblock \emph{Transactions on Machine Learning Research}, 2023.
\newblock ISSN 2835-8856.
\newblock URL \url{https://openreview.net/forum?id=iEq6lhG4O3}.

\bibitem[Wang et~al.(2022)Wang, Lan, Liu, Ouyang, Qin, Lu, Chen, Zeng, and Yu]{wang2022generalizing}
Jindong Wang, Cuiling Lan, Chang Liu, Yidong Ouyang, Tao Qin, Wang Lu, Yiqiang Chen, Wenjun Zeng, and Philip~S. Yu.
\newblock Generalizing to unseen domains: A survey on domain generalization, 2022.

\bibitem[Wang et~al.(2023)Wang, Zhang, Lei, and Zhang]{wang2023sharpnessaware}
Pengfei Wang, Zhaoxiang Zhang, Zhen Lei, and Lei Zhang.
\newblock Sharpness-aware gradient matching for domain generalization, 2023.

\bibitem[Wang and Jordan(2022)]{wang2022desiderata}
Yixin Wang and Michael~I. Jordan.
\newblock Desiderata for representation learning: A causal perspective, 2022.

\bibitem[Wang et~al.(2020)Wang, Li, and Kot]{wang2020heterogeneous}
Yufei Wang, Haoliang Li, and Alex~C. Kot.
\newblock Heterogeneous domain generalization via domain mixup.
\newblock In \emph{ICASSP 2020 - 2020 IEEE International Conference on Acoustics, Speech and Signal Processing (ICASSP)}, pages 3622--3626, 2020.
\newblock \doi{10.1109/ICASSP40776.2020.9053273}.

\bibitem[Wang and Veitch(2022)]{wang2022cisa}
Zihao Wang and Victor Veitch.
\newblock The causal structure of domain invariant supervised representation learning, 2022.
\newblock URL \url{https://arxiv.org/abs/2208.06987}.

\bibitem[Watson(1964)]{watson1964}
G.~S. Watson.
\newblock Smooth regression analysis.
\newblock \emph{Sankhya: The Indian Journal of Statistics}, Series A\penalty0 (26):\penalty0 359--372, 1964.

\bibitem[Wilson et~al.(2015)Wilson, Hu, Salakhutdinov, and Xing]{wilson2015kernellearning}
Andrew~Gordon Wilson, Zhiting Hu, Ruslan Salakhutdinov, and Eric~P. Xing.
\newblock Deep kernel learning, 2015.
\newblock URL \url{https://arxiv.org/abs/1511.02222}.

\bibitem[Xu et~al.(2019)Xu, Zhang, Ni, Li, Wang, Tian, and Zhang]{xu2019adversarial}
Minghao Xu, Jian Zhang, Bingbing Ni, Teng Li, Chengjie Wang, Qi~Tian, and Wenjun Zhang.
\newblock Adversarial domain adaptation with domain mixup, 2019.

\bibitem[Yao et~al.(2022)Yao, Wang, Li, Zhang, Liang, Zou, and Finn]{yao2022lisa}
Huaxiu Yao, Yu~Wang, Sai Li, Linjun Zhang, Weixin Liang, James Zou, and Chelsea Finn.
\newblock Improving out-of-distribution robustness via selective augmentation, 2022.

\bibitem[Yue et~al.(2022)Yue, Zhang, Zhao, Sangiovanni-Vincentelli, Keutzer, and Gong]{yue2022domain}
Xiangyu Yue, Yang Zhang, Sicheng Zhao, Alberto Sangiovanni-Vincentelli, Kurt Keutzer, and Boqing Gong.
\newblock Domain randomization and pyramid consistency: Simulation-to-real generalization without accessing target domain data, 2022.

\bibitem[Zhang et~al.(2021)Zhang, Zhang, Ye, Fang, and Han]{zhang2021hybrid}
Dequan Zhang, Ning Zhang, Nan Ye, Jianguang Fang, and Xu~Han.
\newblock Hybrid learning algorithm of radial basis function networks for reliability analysis.
\newblock \emph{IEEE Transactions on Reliability}, 70\penalty0 (3):\penalty0 887--900, 2021.
\newblock \doi{10.1109/TR.2020.3001232}.

\bibitem[Zhang et~al.(2018{\natexlab{a}})Zhang, Cisse, Dauphin, and Lopez-Paz]{zhang2018mixup}
Hongyi Zhang, Moustapha Cisse, Yann~N. Dauphin, and David Lopez-Paz.
\newblock mixup: Beyond empirical risk minimization, 2018{\natexlab{a}}.

\bibitem[Zhang et~al.(2018{\natexlab{b}})Zhang, Yang, Ma, and Wu]{zhang2018decisiontrees}
Quanshi Zhang, Yu~Yang, Haotian Ma, and Ying~Nian Wu.
\newblock Interpreting cnns via decision trees, 2018{\natexlab{b}}.
\newblock URL \url{https://arxiv.org/abs/1802.00121}.

\bibitem[Zhou et~al.(2021)Zhou, Jiang, Shui, Wang, and Chaib-draa]{zhou2021optimaltransport}
Fan Zhou, Zhuqing Jiang, Changjian Shui, Boyu Wang, and Brahim Chaib-draa.
\newblock Domain generalization via optimal transport with metric similarity learning.
\newblock \emph{Neurocomputing}, 456:\penalty0 469--480, oct 2021.
\newblock \doi{10.1016/j.neucom.2020.09.091}.
\newblock URL \url{https://doi.org/10.1016%2Fj.neucom.2020.09.091}.

\bibitem[Zhou et~al.(2020)Zhou, Yang, Hospedales, and Xiang]{zhou2020generation}
Kaiyang Zhou, Yongxin Yang, Timothy Hospedales, and Tao Xiang.
\newblock Deep domain-adversarial image generation for domain generalisation, 2020.

\bibitem[Zhou et~al.(2022)Zhou, Liu, Qiao, Xiang, and Loy]{zhou2022survey}
Kaiyang Zhou, Ziwei Liu, Yu~Qiao, Tao Xiang, and Chen~Change Loy.
\newblock Domain generalization: A survey.
\newblock \emph{{IEEE} Transactions on Pattern Analysis and Machine Intelligence}, pages 1--20, 2022.
\newblock \doi{10.1109/tpami.2022.3195549}.
\newblock URL \url{https://doi.org/10.1109%2Ftpami.2022.3195549}.

\end{thebibliography}
